\documentclass[acmtog,nonacm]{acmart}

\usepackage{booktabs}
\usepackage{cleveref}
\setcopyright{none}

\begin{document}

\title{OmniLatent: Unified Latent for 3D Part Modeling}
\author{Yunhan Yang}
\authornote{Equal Contribution.}
\orcid{0009-0008-0141-1545}
\affiliation{%
  \institution{University of Hong Kong}
  \city{Hong Kong}
  \country{Hong Kong}}
\email{yhyang.myron@gmail.com}

\author{Yunfei Zhao}
\authornotemark[1]
\orcid{0009-0008-4243-1710}
\affiliation{%
  \institution{Tencent Technology}
  \city{Shenzhen}
  \country{China}}
\email{502022230024@smail.nju.edu.cn}

\author{Zeqiang Lai}
\authornotemark[1]
\orcid{0009-0005-6102-4916}
\affiliation{%
  \institution{The Chinese University of Hong Kong}
  \city{Hong Kong}
  \country{Hong Kong}}
\email{laizeqiang@outlook.com}

\author{Qingxiang Lin}
\orcid{0009-0003-8789-6822}
\affiliation{%
  \institution{Tencent Technology}
  \city{Shenzhen}
  \country{China}}
\email{qingxiang.lin@foxmail.com}

\author{Jiachen Xu}
\orcid{0009-0003-4854-885X}
\affiliation{%
  \institution{Tencent Technology}
  \city{Shenzhen}
  \country{China}}
\email{chanjcxu@tencent.com}

\author{Minxiu Wang}
\orcid{0009-0000-0538-4283}
\affiliation{%
  \institution{Nanjing University}
  \city{Nanjing}
  \country{China}}
\email{minxiuwang@smail.nju.edu.cn}

\author{Xintong Han}
\orcid{0000-0003-3420-2287}
\affiliation{%
  \institution{Tencent Technology}
  \city{Shenzhen}
  \country{China}}
\email{xintong@umd.edu}

\author{Xiangyu Yue}
\orcid{0000-0002-6887-2046}
\affiliation{%
  \institution{The Chinese University of Hong Kong}
  \city{Hong Kong}
  \country{Hong Kong}}
\email{xyyue@ie.cuhk.edu.hk}

\author{Xihui Liu}
\authornote{Corresponding Authors.}
\orcid{0000-0003-1831-9952}
\affiliation{%
  \institution{University of Hong Kong}
  \city{Hong Kong}
  \country{Hong Kong}}
\email{xihuiliu@eee.hku.hk}

\author{Chunchao Guo}
\authornotemark[2]
\orcid{0009-0001-7465-802X}
\affiliation{%
  \institution{Tencent Technology}
  \city{Shenzhen}
  \country{China}}
\email{chunchaoguo@gmail.com}

\renewcommand{\shortauthors}{Yang et al.}

\begin{abstract}
3D part decomposition is a central capability for editable asset creation, yet existing workflows typically segment a completed mesh as a post-processing step, making the result sensitive to geometric artifacts, texture ambiguity, and limited part-level annotations. We present OmniLatent, a unified latent modeling framework for 3D part decomposition. Instead of predicting part labels in isolation, OmniLatent learns a shared latent space that jointly encodes geometry, deformation, appearance, and binary part IDs, allowing data-rich geometric and visual priors to support data-scarce part learning. Given a single input image, the base model generates geometry, deformation, texture, and surface part IDs in one pass, producing a part-aware 3D asset without a separate segmentation stage. We further tune the pretrained model for two extensions: a point-prompt adapter that maps arbitrary 3D point prompts with binary part IDs into latent tokens for full segmentation, refinement, and part subdivision, and an amodal part completion model that completes partial surface masks into a coherent full-object decomposition through dynamic per-part token allocation, cross-attention to the given geometry latent, and alternating part-local and global attention. Extensive experiments show that OmniLatent produces semantically meaningful part segmentations with precise boundaries, supports interactive part editing, and completes partial part observations while preserving global object consistency.
\end{abstract}


\begin{teaserfigure}
  \includegraphics[width=\textwidth]{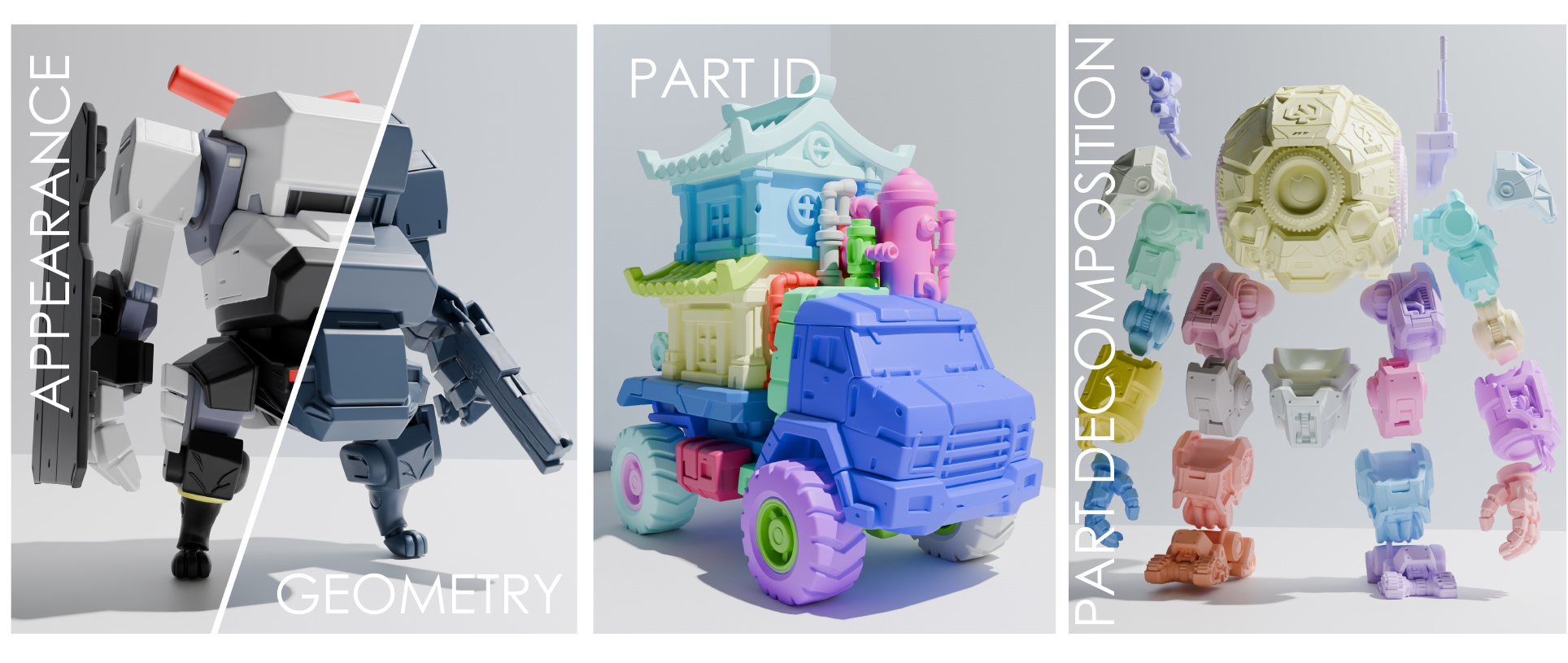}
  \caption{
  OmniLatent generates geometry, deformation, appearance, and coherent surface part IDs from a single image in a unified latent space.}
  \label{fig:teaser}
\end{teaserfigure}

\maketitle

\section{Introduction}
The creation of structured, editable 3D assets is a cornerstone of modern visual computing, supporting applications from game production and cinematic content creation to digital twins, robotic simulation, and interactive design. Recent image-conditioned 3D generative models~\cite{zhao2025hunyuan3d,hunyuan3d2025hunyuan3d,lei2025hunyuan3d,zhang2024clay,li2025triposg} can synthesize increasingly detailed geometry and appearance from a single image. However, their outputs are still mostly monolithic surfaces: the object may look plausible, but its semantic components are not explicitly separated. This missing part structure limits downstream editing, material assignment, animation, physical fabrication, and analysis, all of which require an object to be decomposed into coherent, addressable parts.

A common solution is to first generate or reconstruct a full 3D asset and then run a separate part segmentation model~\cite{ma2025p3,yang2024sampart3d,liu2025partfield}. This post-hoc paradigm is convenient, but it decouples part reasoning from the generative representation that produced the asset. As a result, segmentation quality depends heavily on the input mesh quality and often degrades near thin structures, sharp boundaries, or texture-defined semantic transitions. It also makes part granularity difficult to control: a post-hoc segmenter may split visually coherent components, merge distinct functional parts, or produce boundaries that do not match the generated surface. For part-aware asset creation, the central problem is therefore not only to synthesize a plausible 3D object, but to generate a surface whose semantic part layout is coherent, editable, and aligned with geometry and appearance.

Part decomposition poses an additional data challenge. Holistic geometry and texture data are relatively abundant, while high-quality 3D part annotations are much scarcer. Learning part segmentation directly from part labels alone therefore underuses the strong geometric and visual priors available in larger 3D asset collections~\cite{li2026segvigen}. Practical part cues are also often incomplete: single-view masks or sparse user interactions may only cover visible surface regions, while downstream editing and animation require complete object-level parts. Although 2D foundation models~\cite{zhao2025diception,le2025one} can provide useful masks or prompts, directly transferring their predictions to 3D remains ambiguous due to occlusion, viewpoint bias, and imperfect correspondence between image regions and 3D surfaces.

These limitations suggest that 3D part decomposition should not be treated as an isolated labeling problem. Instead, it should be grounded in a representation that already understands the object's geometry and appearance. This motivates our core question: \textit{Can a unified latent space provide the geometric, visual, and semantic context needed for robust 3D part decomposition?} Prior unified 3D representations have mainly focused on representation formats or holistic asset generation. For example, TRELLIS~\cite{xiang2024structured} unifies different 3D output formats, UniLat3D~\cite{wu2025unilat3d} simplifies textured mesh generation, and UniPart~\cite{he2025unipart} explores geometry-part integration. In contrast, our goal is to use unified latent modeling as the foundation for 3D part decomposition, covering surface part segmentation and amodal part completion. Geometry, deformation, and appearance are generated together with part IDs because they provide the structural and visual evidence needed for reliable part boundaries.

We present OmniLatent, a unified latent modeling framework for 3D part decomposition. As illustrated in \Cref{fig:teaser}, OmniLatent builds upon a voxel-query-based representation~\cite{lai2025lattice} that encodes geometry, deformation, appearance, and binary part IDs into a shared \textbf{OmniLatent space}. The key design principle is to let data-rich modalities, especially geometry and appearance, shape a strong latent prior that can be transferred to data-scarce part decomposition. We first train a unified VAE that compresses heterogeneous surface-aligned fields into the same latent token space with a shared encoder and attribute-specific decoders. We then train an MM-DiT latent generator with rectified flow to synthesize this latent representation from a single image.

The base OmniLatent model is trained progressively. Large-scale geometry data first establishes a structural latent space, textured assets then align appearance with geometry, and a smaller set of part-aware assets finally teaches the latent space to decode coherent part IDs. This curriculum allows part learning to benefit from geometric and visual priors without requiring part annotations at the same scale as geometry or texture. At inference, the base model takes a single image and generates geometry, deformation, texture, and surface part IDs in one pass. When 2D segmentation masks are available, they can optionally guide the desired part granularity; when they are absent, the same model still produces autonomous surface part segmentation from the single image.

Beyond fully automatic segmentation, practical part editing requires interactive control~\cite{tsui2026point2part}. We therefore tune the pretrained model with a point-prompted latent adaptation that represents each prompt as a 3D coordinate with a 6-bit binary part ID. A lightweight MLP maps an arbitrary number of such prompts into latent tokens, which are appended to the denoising sequence. This design avoids a positive-negative prompt formulation: different part IDs directly define distinct semantic groups, enabling full segmentation from sparse prompts, refinement of an existing segmentation, and subdivision of a coarse region into finer parts. Because the adapter is initialized from the pretrained latent generator, it reuses the same geometry, appearance, and part priors instead of learning an interactive segmentation model from scratch.

We further extend OmniLatent to \textit{amodal part completion}: given full object geometry and partial surface masks for all parts, the model completes them into a coherent full-object decomposition. The completion model allocates latent tokens dynamically according to the number of visible voxels assigned to each part, allowing larger parts to receive more representational capacity. For each part, RoPE-anchored tokens preserve observed surface evidence, while additional no-RoPE tokens flexibly represent unobserved regions. A geometry latent encoded by the VAE is injected through cross-attention, and alternating part-local and global attention layers balance fine-grained part completion with full-object consistency.

Our main contributions are summarized as follows:

\begin{itemize}
\item We propose OmniLatent, a unified latent modeling framework that generates geometry, deformation, texture, and surface part IDs from a single image in one pass, allowing geometric and visual priors to support 3D part decomposition with or without 2D segmentation mask guidance.
\item We tune the pretrained model for flexible part control, including point-prompted full segmentation, segmentation refinement, and part subdivision with arbitrary 3D point prompts encoded by binary part IDs.
\item We further adapt the pretrained latent model to amodal part completion with dynamic per-part token allocation and a RoPE/no-RoPE token design, preserving observed surface evidence while flexibly completing unobserved regions into coherent parts.
\item Extensive experiments validate OmniLatent across single-image generation, controllable segmentation, prompt-based editing, and full-object part completion, demonstrating accurate part boundaries, effective user control, globally consistent completion, and competitive supporting geometry and appearance quality.
\end{itemize}

\section{Related Work}
\subsection{3D Shape Generation}
The field of 3D shape generation has evolved from lifting 2D diffusion priors to modeling native 3D distributions directly. Early influential works, such as DreamFusion~\cite{poole2022dreamfusion}, leverage Score Distillation Sampling (SDS) to optimize 3D representations using feedback from 2D diffusion models~\cite{rombach2022high}. 
Zero-1-to-3~\cite{liu2023zero1to3} pioneered image-conditioned novel view synthesis, providing a critical foundation for multi-view pipelines~\cite{liu2023syncdreamer, long2024wonder3d, shi2023zero123plus, liu2023one2345, zou2024triplane}, which focus on generating consistent image sets to reconstruct geometry.
To achieve more robust synthesis, recent research has gravitated toward native 3D latent diffusion. 
This paradigm utilizes Variational Autoencoders (VAEs)~\cite{dai2017shape, wu2024direct3d, zhao2024michelangelo, zhang2024clay, li2024craftsman, lan2025ln3diff, zhang2025compress3d, hui2022neural, koo2023salad, chou2023diffusion, shim2023diffusion, li2023diffusion, li2025triposg, huang2025midi} to compress complex 3D shapes into compact, geometry-aware latent spaces for effective generation. 
Notable examples include 3DShape2VecSet~\cite{zhang20233dshape2vecset} for set-structured data, CLAY \cite{zhang2024clay} for large-scale hierarchical modeling, TRELLIS~\cite{xiang2024structured} for unified representation formats, and 3DTopia-XL~\cite{chen20253dtopia} for primitive-based high-quality asset generation.
Complementary to diffusion, autoregressive models like MeshGPT~\cite{siddiqui2024meshgpt} and its successors~\cite{chen2024meshxl,chen2024meshanything,chen2024meshanythingv2,hao2024meshtron,weng2024scaling,tang2024edgerunner,liu2025freemesh,wang2025nautilus,zhao2025deepmesh} treat mesh generation as a sequential prediction task. 
While these methods have achieved high fidelity in holistic shape synthesis, they often struggle to model continuous multi-attribute details, such as precise geometric sharp edges, fine-grained textures, or parts, which remains an open challenge that we aim to address.

\subsection{3D Texture Generation}
3D texture synthesis has evolved through three primary paradigms: optimization-based distillation, projection-based multi-view synthesis, and native 3D generation. Early influential works~\cite{chen2023fantasia3d,lin2023magic3d,qian2023magic123,wang2023prolificdreamer} utilize Score Distillation Sampling (SDS)~\cite{poole2022dreamfusion} to iteratively optimize 3D appearance from 2D diffusion priors. While capable of producing detailed textures, these optimization-based methods are computationally expensive and frequently suffer from the ``Janus problem''. To address efficiency, another line of research~\cite{huang2025mv,he2025materialmvp,zeng2024paint3d,zhao2025hunyuan3d,zhang2024clay} employs a projection-based paradigm, where consistent multi-view images are generated and then ``painted'' back onto the 3D surface. Related refinement pipelines further enhance low-quality 3D inputs through coupled texture and geometry updates~\cite{ryu2025elevating}. Although Multi-view Diffusion models~\cite{liu2023syncdreamer, long2024wonder3d} improve global consistency by conditioning on geometric maps, the inherent limitations of the painting process often lead to visible seams and misalignments with fine-grained geometric features. To overcome these artifacts, native 3D texture generation seeks to synthesize appearance directly within 3D space. While recent native 3D texture generation frameworks~\cite{lai2025natex,chen2025lafite,liu2024texoct,yu2023texture,yu2024texgen,yuan2025seqtex} have explored predicting colors for points or UV maps by conditioning on fixed geometry, such decoupled generation often struggles to maintain strict coherence between the synthesized texture and the underlying structure.

\subsection{3D Part Decomposition and Generation}
Beyond global shape synthesis, recent research has increasingly prioritized part-aware 3D assets to achieve fine-grained control and structural editability~\cite{zhao2026flexpart,zhu2026partllm,su2026s}. One major direction partitions existing 3D surfaces into semantic parts~\cite{ma2025p3,yang2024sampart3d,liu2025partfield,deng2025geosam2,zhu2025partsam,paul2025name}. These methods are effective for organizing completed meshes~\cite{hadgi2026patchalign3d,jin2026cosmo3d}, but their outputs are inherently tied to the quality of the input surface and the available segmentation cues. Complex boundaries, thin structures, and texture-dependent semantic transitions remain difficult when part prediction is decoupled from the representation that models geometry and appearance.

Another line of research focuses on the direct synthesis of part-aware 3D assets~\cite{yan2025x,li2025moca,yang2025omnipart,lin2025partcrafter,tang2025efficient,dong2025one,ding2025fullpart,he2025unipart,yang2025partdiffuser,zhang2025bang}. Early attempts in this category utilized 2D part-mask supervision to guide multi-view reconstruction~\cite{liu2024part123, chen2024partgen}, though they frequently inherited the view-inconsistency issues of their underlying 2D priors. More recent methods introduce structured latent frameworks or part-specific representations to improve controllability and coherence~\cite{jin2025uniart,zhao2026seeing,gu2026unified,lin2026seg3dparts}. In contrast, OmniLatent focuses on surface part segmentation within a unified latent space that also models geometry, deformation, and appearance. This allows part prediction, prompt-based refinement, and amodal part completion to reuse the same aligned geometric and visual priors rather than treating parts as isolated clusters or post-hoc labels.

\begin{figure*}
\centering
\includegraphics[width=\linewidth]{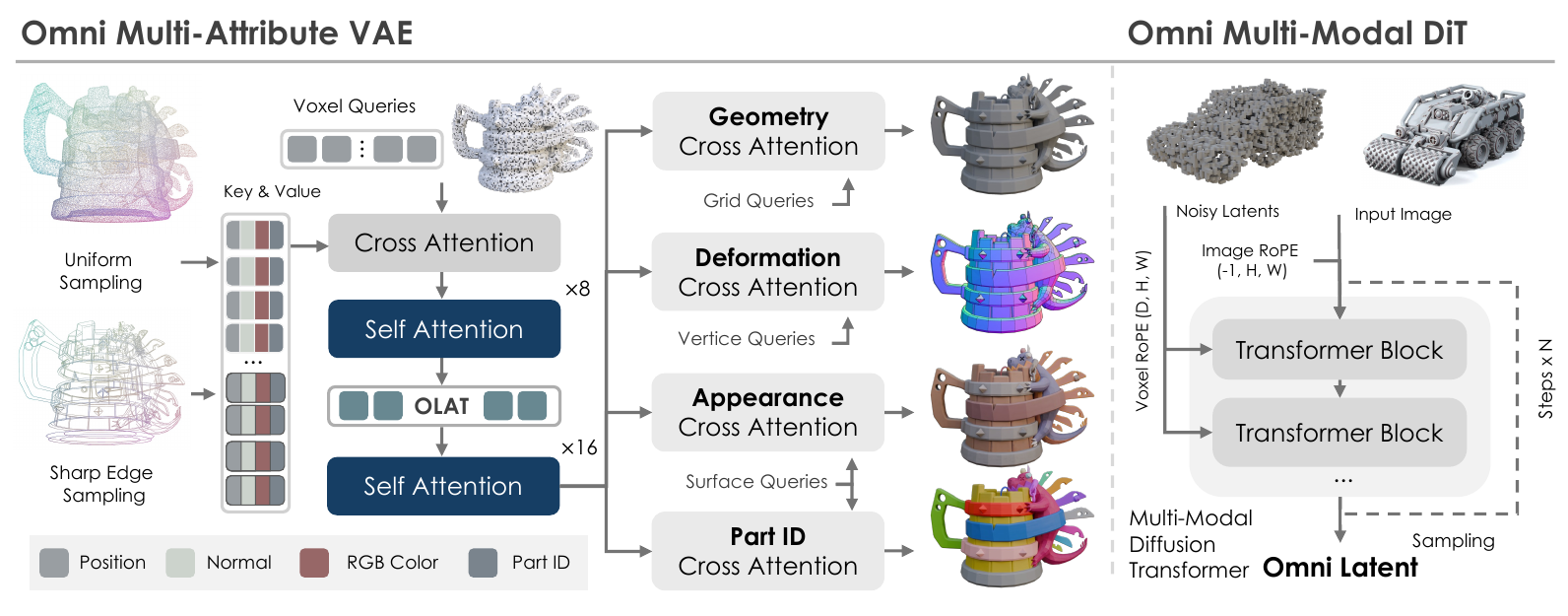}
\caption{\textbf{Unified Latent Modeling and Pretraining}. OmniLatent performs 3D part decomposition through unified latent modeling. Geometry, deformation, appearance, and binary part IDs are encoded into a shared OmniLatent space, then generated by an MM-DiT backbone from an input image with optional 2D mask conditions. The aligned latent space allows geometric and visual priors to support surface part segmentation, and the pretrained model can further be adapted for point-prompted control and amodal part completion. The left branch shows the cascaded VAE decoder: SDF reconstruction defines the surface, whose vertices subsequently query deformation, appearance, and Part IDs from the same latent tokens.}
\label{fig:pipeline}
\end{figure*}

\section{Unified Latent Modeling for 3D Part Decomposition}
As shown in \Cref{fig:pipeline}, OmniLatent casts 3D part decomposition as a unified latent modeling problem. Rather than learning part labels independently from geometry and appearance, we encode geometry, deformation, appearance, and part IDs into a shared OmniLatent space. This design has two goals. First, part predictions should be spatially aligned with the generated surface and appearance, so that boundaries can follow geometric and visual evidence. Second, scarce part annotations should benefit from the stronger priors learned from large-scale geometry and texture data. We first define a surface-aligned multi-attribute representation (\Cref{sec:3-1}), then introduce the unified VAE (\Cref{sec:3-2}), the image- and mask-conditioned latent generator (\Cref{sec:3-3}), point-prompted segmentation and refinement (\Cref{sec:3-4}), and amodal part completion (\Cref{sec:3-5}). The base generator handles single-image part-aware asset generation with optional 2D mask guidance, while the point-prompted and completion models are adapted from the pretrained model and reuse the same OmniLatent representation.

\subsection{Surface-Aligned Multi-Attribute Representation}
\label{sec:3-1}

OmniLatent builds upon VoxSet~\cite{lai2025lattice} architecture and representation, which inherits the set-structured cross-attention mechanism of 3DShape2VecSet~\cite{zhang20233dshape2vecset}. Voxel queries provide an ordered spatial scaffold that can be evaluated at different resolutions, making them suitable for aligning surface attributes with latent tokens. Here, ``ordered'' refers to the spatial order induced by the voxel-query scaffold: each token corresponds to a spatial query location, while its sequence index has no independent semantic meaning. Our extension turns this scaffold into a multi-attribute representation where geometry, deformation, appearance, and part IDs are defined over the same object surface and compressed into a shared latent space.

For a 3D object, we construct a VAE input point set $\mathbf{X} \in \mathbb{R}^{N \times 16}$, where each point contains three aligned fields: a geometric field $\mathbf{P} \in \mathbb{R}^{N \times 7}$, an appearance field $\mathbf{C} \in \mathbb{R}^{N \times 3}$, and a part field $\mathbf{Q} \in \{0,1\}^{N \times 6}$. The concatenated field is written as $\mathbf{X}=[\mathbf{P},\mathbf{C},\mathbf{Q}]$.

\paragraph{Geometry field}
The geometric foundation of the asset is represented by $\mathbf{P}$, which contains 3D coordinates, surface normals, and a binary sharpness indicator. These samples supervise an SDF decoder used for surface extraction with Marching Cubes~\cite{lorensen1998marching}. Following recent high-quality 3D asset generation practice~\cite{zhao2025hunyuan3d}, we combine uniform surface sampling with additional samples near sharp edges, so that the latent space observes both global shape and high-frequency structural cues.

\paragraph{Appearance field}
The appearance field $\mathbf{C}$ stores RGB values sampled on the object surface. During watertight mesh processing, we preserve original face colors whenever possible; for adjusted faces, we assign colors through nearest-neighbor interpolation from the original surface. Since the processed faces are small, this strategy keeps texture supervision aligned with the reconstructed surface without introducing strong color noise.

\paragraph{Part field}
The part field $\mathbf{Q}$ stores a 6-bit binary encoding of surface part IDs. This representation supports up to $2^6=64$ parts, which is sufficient for the fine-grained objects used in our part-aware training set. The codes are object-local rather than members of a global semantic taxonomy: during training, we randomly assign every part of each object a distinct 6-bit ID, so no cross-object correspondence between numerical IDs is required. Compared with RGB-coded part masks, binary part IDs are more stable under denoising: small continuous color deviations in RGB space can split one semantic part into multiple inconsistent IDs, while bitwise decoding preserves categorical precision.

\paragraph{Deformation field}
We include a lightweight deformation head as part of the geometric output. It is decoded after SDF reconstruction and Marching Cubes, and predicts small vertex offsets that sharpen local surface structures. Its supervision is derived from the same geometric field $\mathbf{P}$, whose positions, normals, and sharpness cues identify vertex-to-surface displacements near edges and corners. This field improves geometric presentation quality, while the part capability itself is learned through the unified latent part field.

\subsection{OmniLatent Autoencoding}
\label{sec:3-2}

Given the concatenated multi-attribute input $\mathbf{X}$, we train a VAE that maps it into the OmniLatent space $\mathcal{Z}$. A shared encoder $E$ compresses $\mathbf{X}$ into an ordered set of latent tokens $\mathbf{z} \in \mathbb{R}^{L \times D}$ via cross-attention. The decoder uses a shared transformer backbone followed by attribute-specific heads $\{G_{sdf}, G_{def}, G_{app}, G_{part}\}$ for SDF, deformation, appearance, and part IDs.

Decoding follows a cascaded surface-query strategy, as visualized in the left branch of \Cref{fig:pipeline}. We first reconstruct the SDF and extract the surface; the resulting surface vertices are then used to query deformation, appearance, and part ID fields from the same latent tokens. This shared query path is important for part segmentation: the part field is decoded on the same surface that defines geometry and appearance, reducing the mismatch between part boundaries and object structure. Decoded surface labels are converted to per-face masks; therefore, the Part-ID output contains no overlapping internal interfaces. Complete internal or occluded part geometry is recovered only by the subsequent amodal completion model.

The VAE is trained with a multi-task objective,
$$\mathcal{L}_{vae} = \lambda_{kl}\mathcal{L}_{KL} + \lambda_{sdf}\mathcal{L}_{sdf} +  \lambda_{def}\mathcal{L}_{def} + \lambda_{app}\mathcal{L}_{app} + \lambda_{part}\mathcal{L}_{part},$$
where $\lambda_i$ are balancing coefficients. Specifically, $\mathcal{L}_{KL}$ regularizes the posterior latent distribution toward the unit Gaussian prior; $\mathcal{L}_{sdf}$ supervises reconstructed signed-distance values at spatial queries; $\mathcal{L}_{def}$ regresses the target offsets of extracted surface vertices; $\mathcal{L}_{app}$ reconstructs surface RGB values; and $\mathcal{L}_{part}$ supervises the six binary Part-ID channels on the object surface. We use a progressive supervision schedule: SDF training first establishes a stable geometric latent space, and deformation, appearance, and part heads are then added with surface-near supervision. This schedule encourages the part decoder to learn on top of an already aligned geometric and visual latent representation.

\begin{figure*}
\centering
\includegraphics[width=\linewidth]{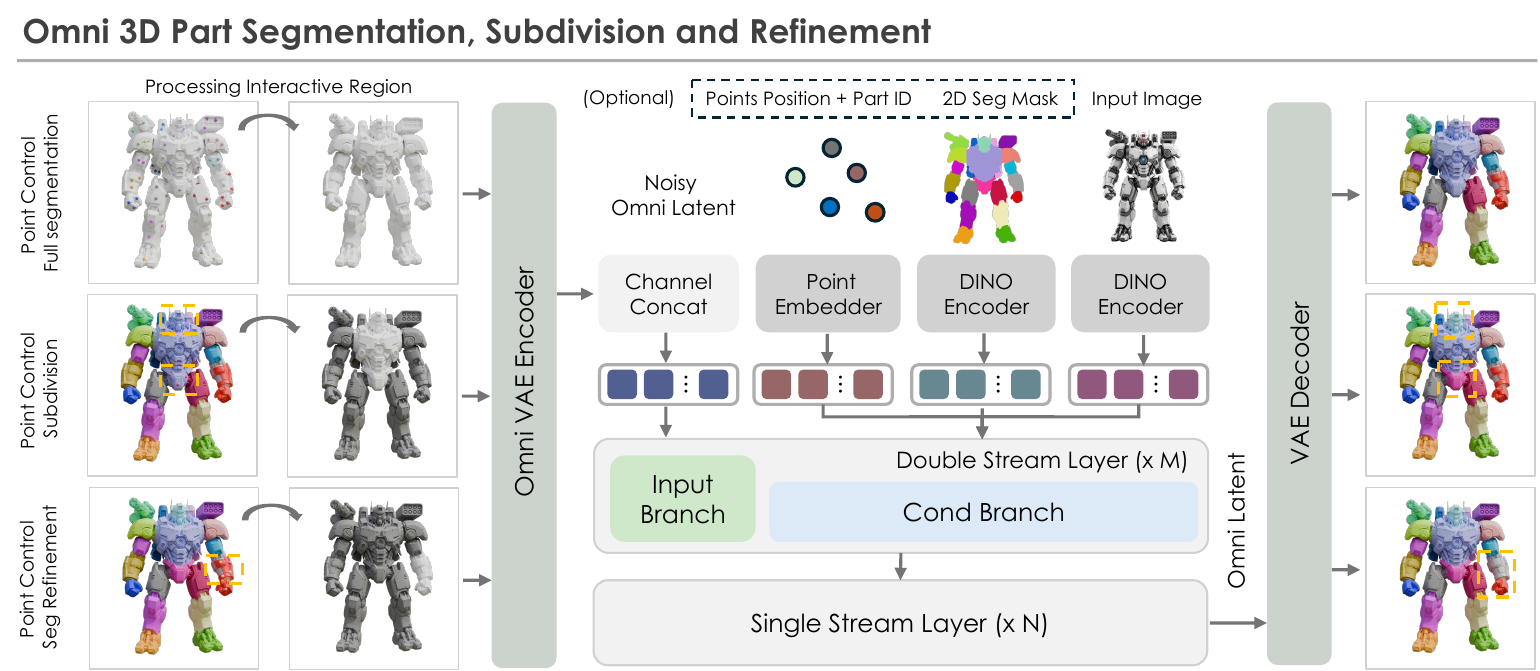}
\caption{\textbf{Unified part segmentation and refinement} using images, 2D masks, and point prompts. We adopt an MM-DiT architecture that integrates multiple control signals: interactive regions and geometric conditions are injected via channel concatenation with the noisy latent, while other conditions are incorporated through sequence concatenation.}
\label{fig:arch_part_seg}
\end{figure*}

\subsection{Image- and Mask-Conditioned Latent Generation}
\label{sec:3-3}

We train an MM-DiT~\cite{esser2024scaling} with rectified flow to generate latent tokens in the OmniLatent space. Let $z_1 \in \mathcal{Z}$ denote a target latent sample and $z_0 \sim \mathcal{N}(0,\mathbf{I})$ denote noise. The interpolation path is $z_t = t z_1 + (1-t) z_0$, and the model $v_\theta$ is trained with
$$\mathcal{L}_{flow} = \mathbb{E}_{t, z_0, z_1} \| v_\theta(z_t, t, \mathbf{c}) - (z_1 - z_0) \|^2,$$
where $\mathbf{c}$ is a sequence of condition tokens. For the base generator, $\mathbf{c}$ always includes the input RGB image and can additionally include a 2D segmentation mask that specifies the desired part granularity. The generated latent tokens are decoded into geometry, deformation, appearance, and part IDs together, so surface part segmentation is learned as part of the same conditional generation process rather than as a separate post-processing step. In the mask-free setting, the model produces autonomous surface part segmentation from the RGB image alone; in the mask-guided setting, the 2D segmentation mask provides controllable part layout cues.

Data heterogeneity is a central challenge: geometry is abundant, textured assets are less abundant, and part-aware 3D objects are much scarcer. We therefore use progressive training. The model first learns large-scale geometry and deformation priors, then incorporates textured assets to align appearance with geometry, and finally fine-tunes on part-aware objects. For samples without appearance or part annotations, the missing fields are zeroed before encoding so that the latent space remains compatible across training stages. In the final stage, the model learns to decode binary part IDs from a latent space already shaped by geometry and appearance.

To handle variable conditions, we use Condition-Aware 3D Rotary Position Embedding (CA-RoPE)~\cite{su2024roformer}. For image-like conditions, the first RoPE dimension indexes the condition type, while the remaining two dimensions encode spatial height and width. In practice, RGB images, optional 2D segmentation masks, and other image-space conditions are concatenated as condition tokens, while their condition-type indices keep them distinguishable from each other and from 3D latent tokens. During part-aware training, 2D segmentation mask conditions are randomly dropped, enabling the base model to operate both with mask-guided granularity control and in the mask-free setting where it generates geometry, deformation, appearance, and surface part IDs from the RGB image alone.

We also apply mask-condition augmentation to bridge the gap between 2D cues and 3D part layouts. Since 2D segmentation masks may be incomplete, view-dependent, or inconsistent with occluded 3D structures, we augment them through random dropping, merging, splitting, component decomposition, internal noise, and erosion. These augmentations encourage the model to treat 2D masks as granularity guidance while still aligning final part boundaries with the generated 3D geometry and appearance.

\subsection{Point-Prompted Segmentation and Refinement}
\label{sec:3-4}

As summarized in \Cref{fig:arch_part_seg}, OmniLatent supports interactive part control through point prompts. A prompt set is written as $\mathcal{P}=\{(\mathbf{x}_i,\mathbf{b}_i)\}_{i=1}^{n}$, where $\mathbf{x}_i \in \mathbb{R}^3$ is a prompted 3D surface point and $\mathbf{b}_i \in \{0,1\}^{6}$ is the binary part ID assigned to that point. We concatenate each coordinate and part code, then map it through a lightweight MLP into the latent token dimension. The resulting $n$ prompt tokens are appended to the denoising sequence and participate in all transformer layers.

This formulation supports any number of prompts and avoids a positive-negative prompt design. Whereas positive-negative clicks specify foreground and background for a single target, our multi-ID prompts distinguish several parts within one prompt set and support full segmentation, refinement, and subdivision in one pass. Different part IDs directly define different groups, so the model can infer a full surface segmentation by propagating these sparse semantic anchors through the geometry-aware latent space. The point-prompted model is fine-tuned from the fully trained single-image generator by sampling point-ID pairs from ground-truth part segmentations, allowing the prompt behavior to reuse the pretrained geometry, appearance, and part priors. During tuning, full-segmentation examples sample 1--10 prompt points per part; for refinement, we randomly merge 2--4 parts and again sample 1--10 points on each original part inside the merged region, training the model to recover finer part boundaries from coarse initial regions.

The same point-prompted model is trained to refine existing segmentations. In this setting, an initial full segmentation is encoded together with geometry by the unified VAE, and the encoded feature is concatenated with the noisy latent tokens as a geometry-aware condition. We additionally provide a binary region indicator: the part selected for refinement or subdivision is marked white, and the remaining context is marked black. This indicator is a soft condition rather than a hard constraint. The model is therefore allowed to adjust boundaries and nearby assignments when the prompts and geometry suggest a more coherent assignment.

\begin{figure}
\centering
\includegraphics[width=\linewidth]{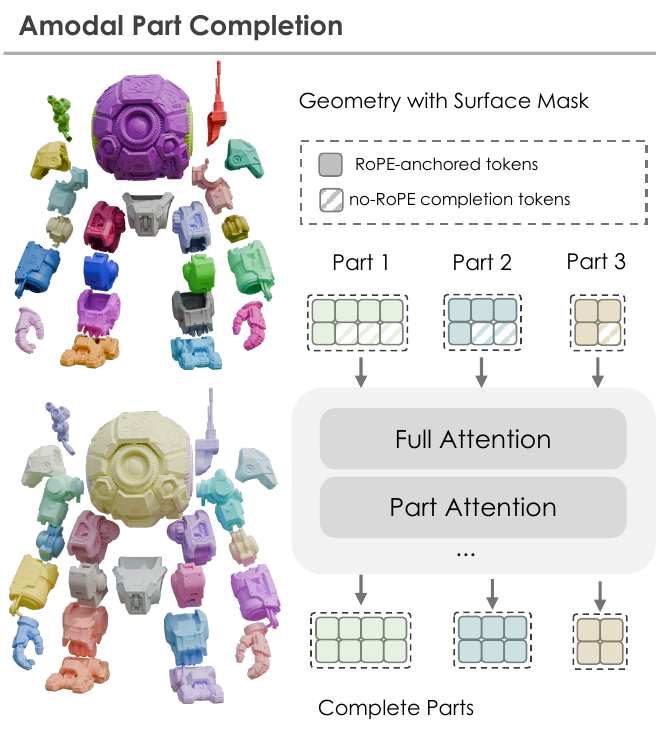}
\caption{\textbf{Amodal part completion.} Given full geometry and partial surface masks for all parts, OmniLatent allocates tokens dynamically according to visible part size. RoPE-anchored tokens preserve observed surface evidence, while no-RoPE completion tokens represent unobserved regions. A geometry latent enters through cross-attention, and alternating part-local and global attention layers complete all parts into a consistent full-object part layout.}
\label{fig:part_completion_method}
\end{figure}

\subsection{Amodal Part Completion}
\label{sec:3-5}

As shown in \Cref{fig:part_completion_method}, we define amodal part completion as completing full object-level parts from partial surface masks on a given geometry. Following the task terminology of HoloPart~\cite{yang2025holopart}, the goal is to recover complete part structures, including unobserved or internal regions, from a holistic mesh and partial surface masks. We adapt the pretrained unified latent model to this task. The input consists of the complete object geometry and partial surface masks for all parts. This geometry is generated by the base OmniLatent model rather than taken from ground truth. The output is a complete part layout over the entire object surface, including both visible and unobserved regions of every part.

The main difficulty is that different parts occupy very different amounts of surface area. Using a fixed number of tokens per part either wastes capacity on small parts or under-represents large parts. We therefore select surface voxels on the given geometry and allocate tokens to each part according to its visible voxel count, with a bounded range around the average per-part token budget to avoid extreme allocations. Specifically, each part token count is clamped to $[0.5, 1.5]$ times the average per-part token budget. Larger visible parts receive more tokens, while small parts keep a compact representation. This dynamic token allocation makes the completion model scale with part size without requiring a fixed part count or fixed per-part length.

For each part, we split its tokens into two groups. The first group is anchored to observed surface regions with RoPE, preserving the spatial identity of the visible mask. The second group contains no-RoPE completion tokens. Since these tokens are not tied to fixed observed positions, they add flexible completion capacity without overwriting the observed evidence, allowing the model to infer the complete extent of the part. To expose the model to all available part identity embeddings, we randomly remap each training part to an ID in $[0,63]$ and inject the corresponding learned part embedding into all tokens of that part. The no-RoPE completion-token ratio is randomly sampled from 20\% to 40\% during training and fixed to 30\% during inference. During training, we know which tokens correspond to observed surface regions and which tokens correspond to regions that must be completed. We therefore upweight the loss on completion tokens by 5, because these unobserved-region tokens are fewer and converge more slowly than visible surface tokens. We also filter training objects in which four or more parts require more than 60\% completion, which avoids highly ambiguous supervision and improves optimization stability.

To keep completion consistent with the entire object, we encode the full geometry with the unified VAE and inject the geometry latent through cross-attention. We also alternate two attention patterns in the completion transformer: part-local attention, where tokens interact within the same part to recover fine-grained structure, and global full attention, where all part tokens interact to maintain full-object consistency. This combination allows each part to be completed independently enough to preserve local detail, while still coordinating with neighboring parts and the overall object shape.

\section{Experiments}
\begin{figure*}[t]
\centering
\includegraphics[width=1\linewidth]{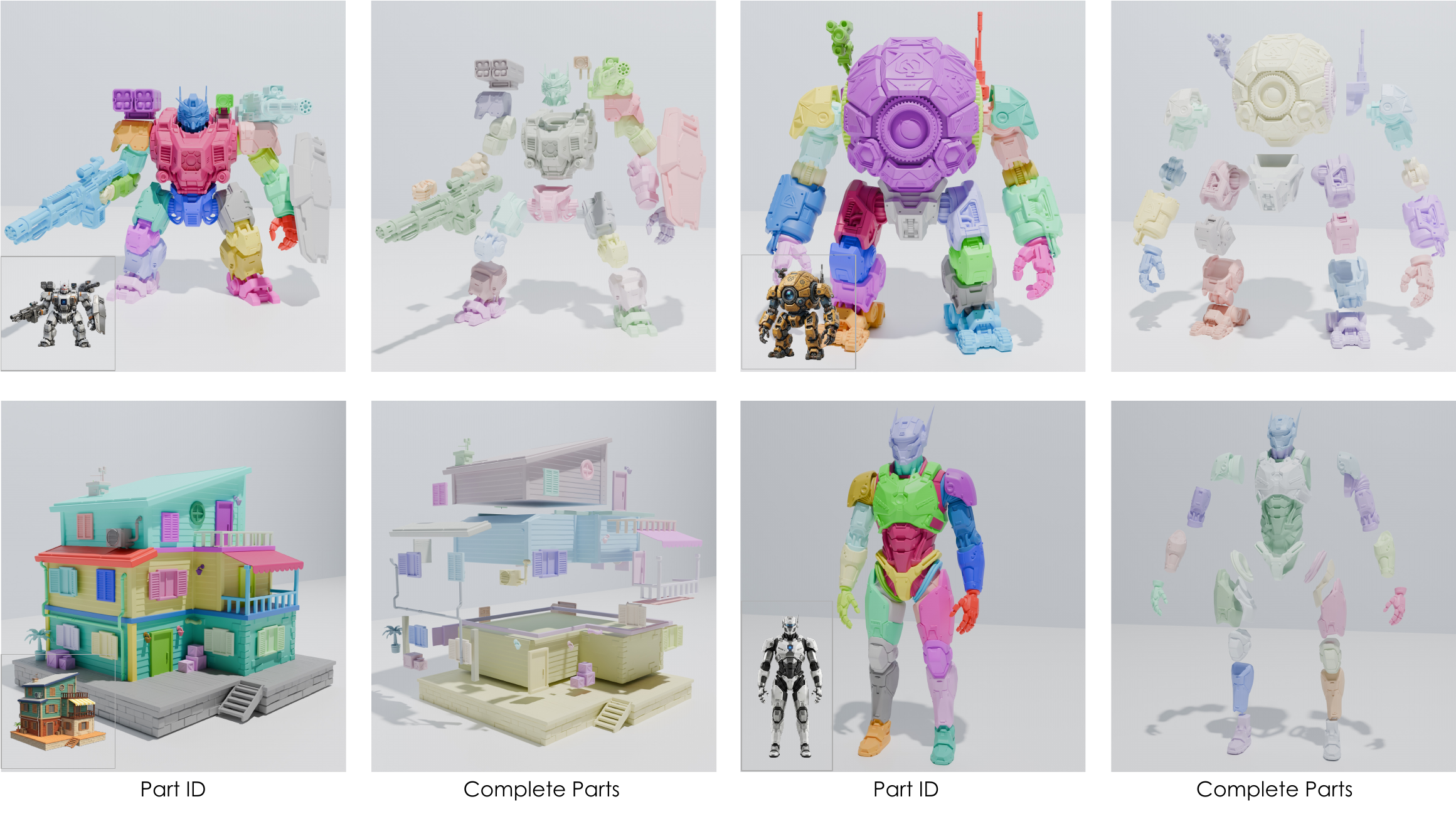}
\caption{\textbf{Surface part segmentation and amodal completion results.} For each example, OmniLatent produces complete surface part segmentation and further completes partial surface masks into complete object-level parts, preserving global structure while recovering missing components.}
\label{fig:part_seg_com}
\end{figure*}

We evaluate OmniLatent as a unified latent framework for 3D part decomposition. Our experiments focus on four part-centered capabilities: single-image joint generation of geometry, deformation, appearance, and surface part IDs; mask-guided controllable segmentation; point-prompted segmentation, refinement, and subdivision after tuning; and amodal part completion after tuning. We additionally report geometry and appearance generation results as supporting evidence that the shared OmniLatent space preserves the quality of the auxiliary modalities used to guide part learning. \Cref{fig:part_seg_com,fig:exp_part_tex_geo} show representative single-image generation and mask-guided segmentation results where the same generated latent sample decodes geometry, appearance, and surface part segmentation.

\begin{figure*}
\centering
\includegraphics[width=\linewidth]{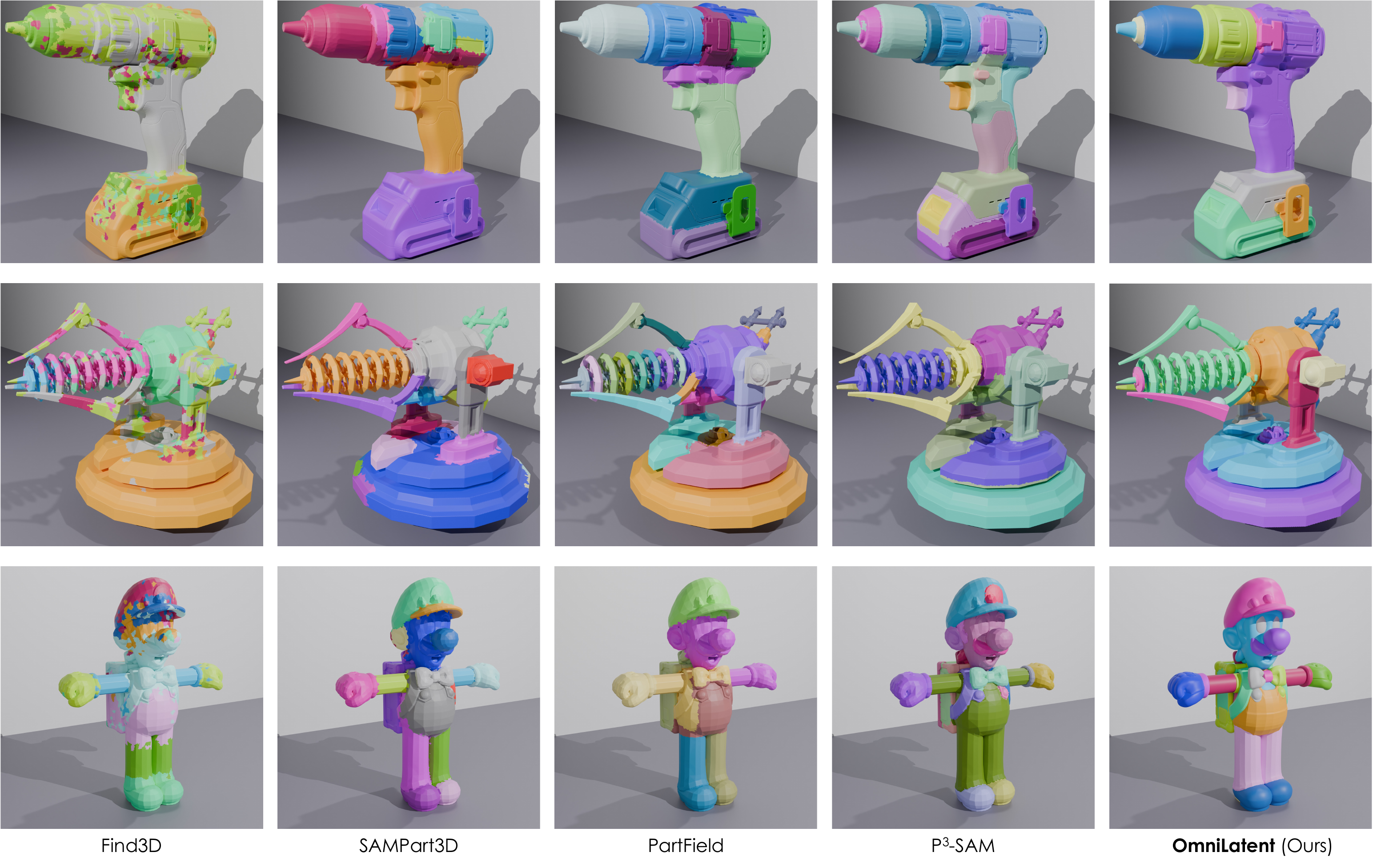}
\caption{\textbf{Qualitative comparison of surface part segmentation} on PartObjaverse-Tiny-wt. OmniLatent produces semantically coherent regions and cleaner boundaries by decoding part IDs from a latent space aligned with geometry and appearance.}
\label{fig:exp_part_com}
\end{figure*}

\subsection{Implementation Details}
\label{sec:4-impl}

\paragraph{Data preparation}
For geometry and deformation, we curate 900K high-quality meshes primarily sourced from Objaverse~\cite{deitke2022objaverse}. For appearance modeling, we use 300K assets that contain both geometry and texture information, targeting albedo to encourage the model to learn lighting-robust appearance. We render conditioning images under diverse illumination by randomly selecting point lights, area lights, and HDRI maps. For part-level modeling, we use 20K part-aware 3D objects. These objects are selected from Objaverse, filtered for meaningful part decomposition, and manually annotated with semantic Part IDs. These assets are processed to ensure consistent geometry and appearance, and the number of semantic parts per object typically ranges from 5 to 30 with an emphasis on semantic rationality and symmetry. We render both RGB conditioning images and corresponding 2D segmentation masks in Blender.

\paragraph{Training and inference}
During unified VAE training, we first optimize the SDF field with a learning rate of $10^{-4}$. We initially train with a token length of 2048 before transitioning to 6144, which improves stability and allows extrapolation to longer sequences during inference. We use random token sampling during VAE training to improve robustness across voxel densities. Once the SDF space is stabilized, we introduce joint supervision for deformation, appearance, and part fields. The loss weights are $\lambda_{kl}=10^{-6}$, $\lambda_{sdf}=1$, $\lambda_{app}=1$, $\lambda_{part}=1$, and $\lambda_{def}=0.1$; the smaller deformation weight prevents structural offsets from dominating the multi-task objective.

Our generative backbone is a 1.9B-parameter MM-DiT trained with a constant learning rate of $5\times10^{-5}$ and a linear warm-up phase. The DiT is trained with a sequence length of 12,248 tokens. To support classifier-free guidance for RGB image and 2D segmentation mask conditions, we randomly replace conditioning embeddings with zero embeddings at a 10\% probability during training, and use a unified CFG scale of 5 at inference. Both the point-prompt and amodal-completion adapters are fully fine-tuned from the pretrained base model on the same 20K part-aware objects for 30K steps with a learning rate of $5\times10^{-5}$. VAE training takes approximately 3 days, while the main DiT training takes approximately 7 days. At inference time, the base OmniLatent model generates geometry, deformation, appearance, and part IDs from a single image in approximately 30 seconds with 8192 latent tokens.

\subsection{Surface Part Segmentation}
\label{sec:4-1}

\paragraph{Settings and metrics}
We evaluate surface part segmentation on PartObjaverse-Tiny~\cite{yang2024sampart3d}, a curated subset of Objaverse~\cite{deitke2023objaverse} with over 3,000 fine-grained part annotations across 200 shapes. Following the evaluation protocol of $\text{P}^3$-SAM~\cite{ma2025p3}, we use the watertight version of the benchmark and report mean Intersection over Union (mIoU). Because OmniLatent decodes part IDs on the generated watertight surface, we assign ground-truth part IDs to the predicted surface by nearest-neighbor transfer. For each ground-truth part, we match the closest predicted part region and compute IoU; the mIoU is averaged over ground-truth parts for each shape and then over all test shapes. Thus, the evaluated part count is determined by the ground-truth annotation rather than by extra predicted fragments.

We consider two segmentation modes. In the autonomous setting, the base model receives only the input image and directly generates geometry, deformation, appearance, and surface part IDs without a 2D segmentation mask. In the controllable setting, a 2D segmentation mask condition is additionally provided to indicate the desired part granularity. The same base model supports both settings because 2D segmentation mask conditions are randomly dropped during training. Unless explicitly stated otherwise, the OmniLatent result of 74.18 mIoU in \Cref{tab:exp_part} uses mask guidance.

\paragraph{Baselines}
We compare with representative surface segmentation methods, including Find3D~\cite{ma2025find}, SAMPart3D~\cite{yang2024sampart3d}, PartField~\cite{liu2025partfield}, and $\text{P}^3$-SAM~\cite{ma2025p3}. These methods operate on existing 3D surfaces and differ in how they control segmentation granularity. Find3D uses predefined label sets to regulate segmentation scale through semantic prompts, while SAMPart3D and PartField provide multi-granularity segmentations through learned features. Following their official configurations, we evaluate SAMPart3D and PartField across multiple granularities and report the best IoU.

\begin{table*}[!ht]
\centering
\caption{Surface part segmentation on PartObjaverse-Tiny-wt, reported in mIoU (\%). OmniLatent is evaluated in the mask-guided setting.}
\resizebox{0.95\linewidth}{!}{

\begin{tabular}{c|c|cccccccc}
\hline
 \textbf{Method}  & \textbf{Avg.} $\uparrow$ &  Human-Shape& Animals& Daily-Used& Buildings & Transportations & Plants& Food & Electronics         \\  \hline  

 Find3D~\cite{ma2025find}   & 20.76 & 25.08 & 24.86 & 21.71 & 14.93 & 12.52 & 22.43 & 27.54 & 17.02  \\ %\cline{3-14} 

 SAMPart3D~\cite{yang2024sampart3d} & 50.02 & 48.49 & 58.66 & 49.00 & 39.52 & 42.68 & 58.37 & 54.94 & 48.48 \\

 PartField~\cite{liu2025partfield} & 52.61 & 54.50 & 60.36 & 56.38 & 38.43 & 44.57 & 55.70 & 58.59 & 52.31 \\

 $\text{P}^3$-SAM~\cite{ma2025p3} & 58.10 & 64.10 & 62.04 & 60.74 & 43.66 & 54.83 & 65.36 & 55.73 & 58.37 \\
  
 \textbf{OmniLatent} (Ours)  & \textbf{74.18} & \textbf{79.36}& \textbf{82.68}& \textbf{76.98}& \textbf{58.94} & \textbf{65.04} & \textbf{78.73} & \textbf{80.72} & \textbf{70.98} \\ \hline
\end{tabular}
}

\label{tab:exp_part}
\end{table*}

\begin{table}[t]
\centering
\caption{Additional surface part segmentation comparisons in mIoU (\%). The same-surface setting evaluates every method on meshes generated by OmniLatent. We also evaluate OmniLatent on the ground-truth watertight surface with and without 2D mask guidance.}
\label{tab:exp_part_fairness}
\setlength{\tabcolsep}{7pt}
\resizebox{\linewidth}{!}{
\begin{tabular}{llc}
\toprule
\textbf{Surface} & \textbf{Method} & \textbf{mIoU}$\uparrow$ \\
\midrule
Generated & Find3D~\cite{ma2025find} & 19.21 \\
Generated & SAMPart3D~\cite{yang2024sampart3d} & 48.73 \\
Generated & PartField~\cite{liu2025partfield} & 51.93 \\
Generated & $\text{P}^3$-SAM~\cite{ma2025p3} & 56.01 \\
Generated & \textbf{OmniLatent} & \textbf{74.18} \\
\midrule
GT watertight & Ours w/o mask & 62.42 \\
GT watertight & \textbf{Ours w/ mask} & \textbf{76.53} \\
\bottomrule
\end{tabular}
}
\end{table}

\paragraph{Results}
As shown in \Cref{tab:exp_part}, OmniLatent achieves higher part segmentation accuracy than the compared baselines. Because the standard comparison evaluates baselines on the ground-truth watertight mesh but OmniLatent on its generated mesh, we additionally evaluate every method on the same OmniLatent-generated surfaces in \Cref{tab:exp_part_fairness}. OmniLatent remains best at 74.18 mIoU, compared with 56.01 for the strongest baseline. On the ground-truth watertight surface, OmniLatent reaches 62.42 mIoU without mask guidance and 76.53 with mask guidance. Ground-truth masks are transferred to generated surfaces by nearest-neighbor assignment, so geometric deviations can lower our generated-surface score. Qualitatively, the unified latent formulation produces coherent part regions with clean boundaries, especially on objects where part transitions are supported by both geometry and appearance. \Cref{fig:exp_part_com} compares OmniLatent with surface segmentation baselines on PartObjaverse-Tiny-wt, while \Cref{fig:part_seg_com} shows complete surface part segmentation results produced by OmniLatent on diverse assets. Unlike post-hoc segmentation pipelines, OmniLatent predicts part IDs from the same latent representation that also decodes the surface and appearance, allowing the segmentation to remain aligned with generated geometry.

\subsection{Point-Prompted Segmentation, Refinement, and Subdivision}
\label{sec:4-2}

\begin{figure*}[t]
\centering
\includegraphics[width=1\linewidth]{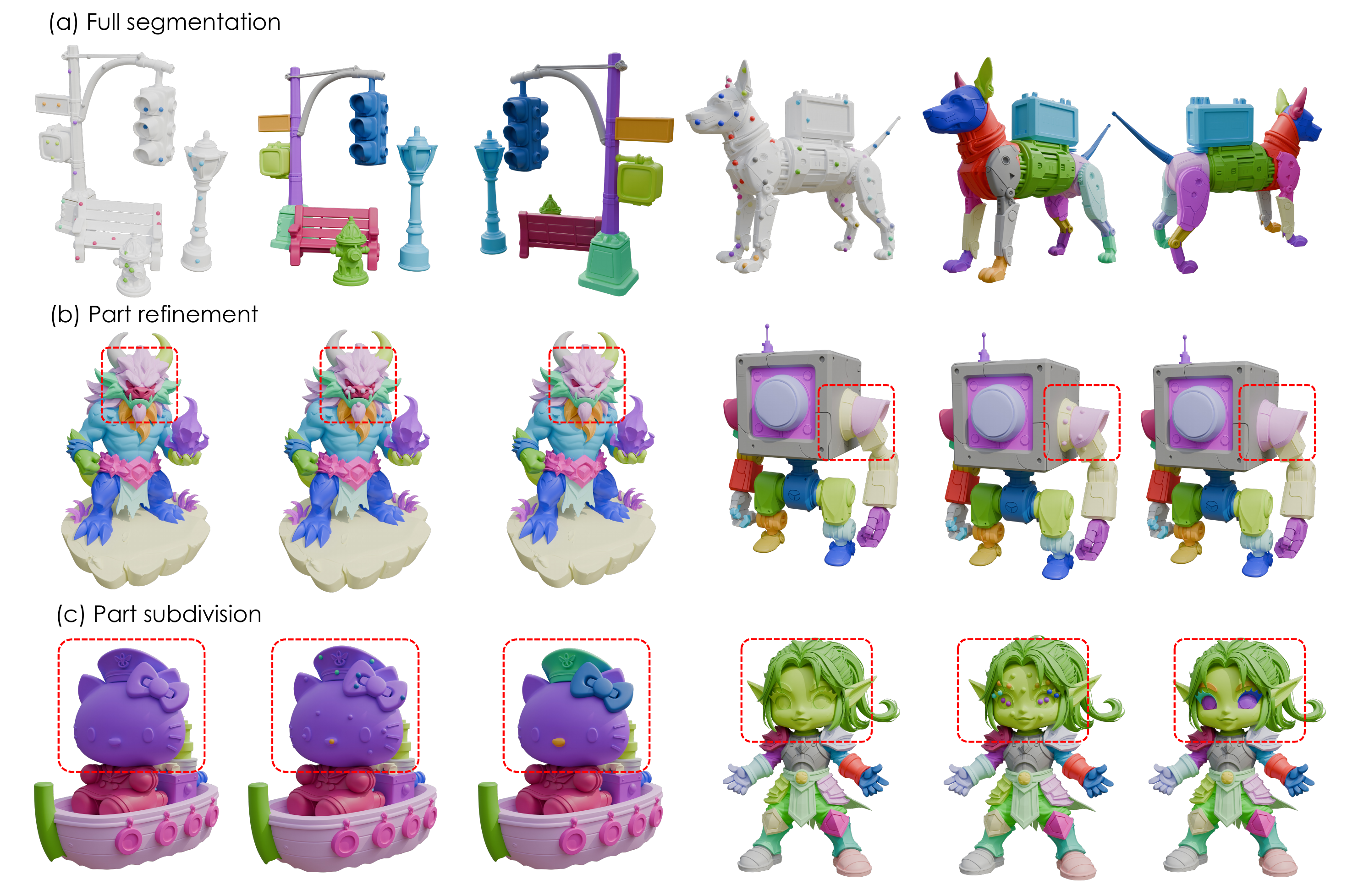}
\caption{\textbf{Point-prompted part control.} OmniLatent supports three prompt-driven editing modes: (a) Full segmentation from sparse point prompts, (b) Part refine for correcting an initialized segmentation, and (c) Part subdivision for splitting a coarse part into finer components.}
\label{fig:part_point_prompt}
\end{figure*}

We evaluate the prompt-adapted model through sparse-prompt control and refinement. For prompt-driven segmentation from scratch, we provide sparse 3D point prompts with binary part IDs and ask the model to infer the full object segmentation. The model accepts any number of prompts, and different part IDs serve as semantic anchors for different regions. For refinement and subdivision, we provide an initialized full segmentation together with point prompts and a binary region indicator that marks the part selected for editing. The indicator is used as conditioning rather than a hard mask, so the model can adjust boundaries when the local geometry or prompt evidence suggests a better assignment. The prompt-adapted model is trained on both sampled full-segmentation prompts and synthetic coarse-to-fine refinement cases, so the same interface covers part creation and correction. For quantitative comparison, we follow the $\text{P}^3$-SAM protocol and use 10 prompts for every method. As reported in \Cref{tab:exp_point_prompt}, OmniLatent achieves 67.32 mIoU for full segmentation and 55.75 mIoU for interactive segmentation, outperforming Point-SAM and $\text{P}^3$-SAM.

\begin{table}[t]
\centering
\caption{Point-prompted surface part segmentation using 10 prompts, reported in mIoU (\%). ``Full Seg.'' evaluates complete segmentation from prompts, while ``Interactive Seg.'' evaluates prompt-guided editing.}
\label{tab:exp_point_prompt}
\setlength{\tabcolsep}{6pt}
\resizebox{\linewidth}{!}{
\begin{tabular}{lcc}
\toprule
\textbf{Method} & \textbf{Full Seg.}$\uparrow$ & \textbf{Interactive Seg.}$\uparrow$ \\
\midrule
Point-SAM~\cite{zhou2024point} & -- & 24.16 \\
$\text{P}^3$-SAM~\cite{ma2025p3} & 58.10 & 49.11 \\
\textbf{OmniLatent} & \textbf{67.32} & \textbf{55.75} \\
\bottomrule
\end{tabular}
}
\end{table}

As shown in \Cref{fig:part_point_prompt}, this prompt interface supports three practical behaviors: (a) Full segmentation from sparse prompts, (b) Part refine for correcting an existing segmentation, and (c) Part subdivision for splitting a coarse region into finer parts. These results demonstrate that the point-prompt adapter turns the pretrained generator into an interactive part editor without changing the underlying OmniLatent space. Since the prompt tokens are appended directly to the denoising sequence, the model can reuse the pretrained geometry, appearance, and part priors while responding to sparse user input.

\subsection{Amodal Part Completion}
\label{sec:4-3}

\begin{figure*}[t]
\centering
\includegraphics[width=\linewidth]{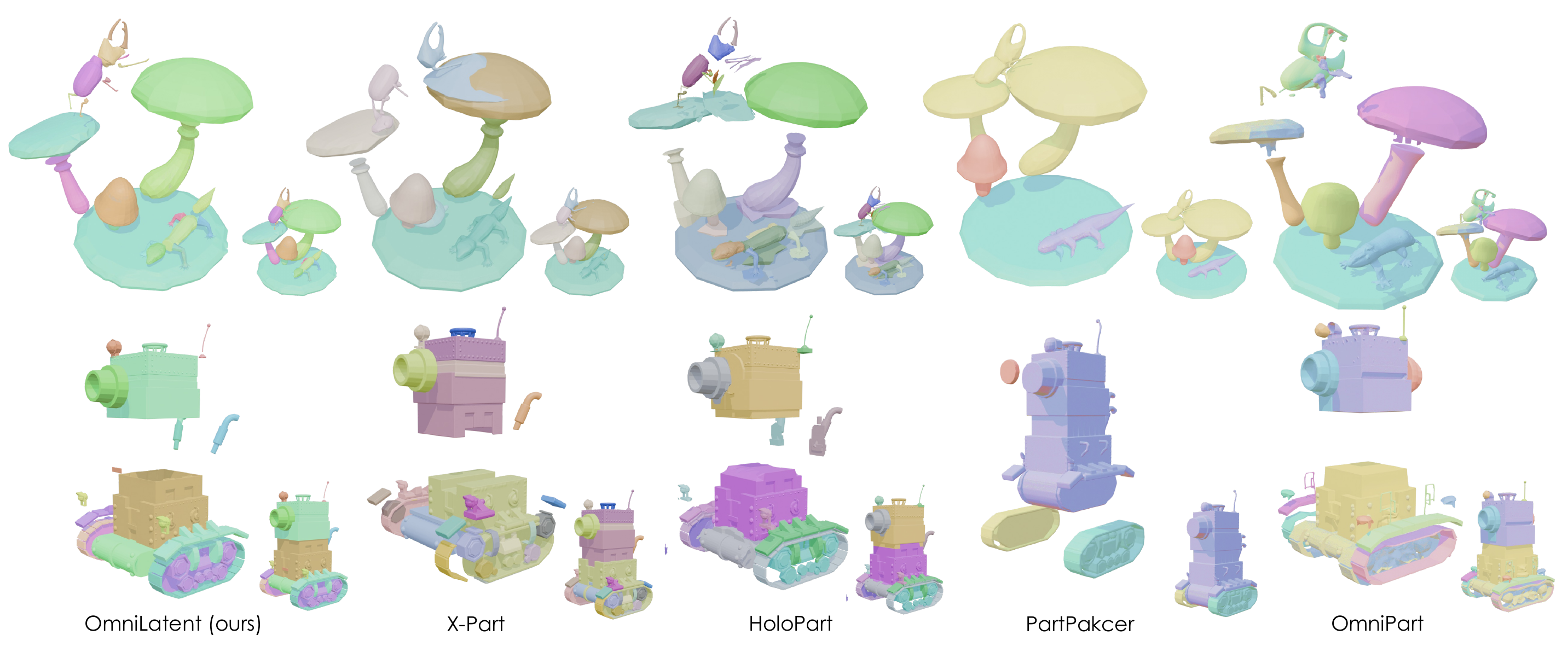}
\caption{\textbf{Qualitative comparison on amodal part completion.} Given full geometry and partial surface masks, OmniLatent preserves visible part evidence and completes coherent full-object parts, producing cleaner part geometry and more consistent object structure than X-Part, HoloPart, PartPacker, and OmniPart.}
\label{fig:part_completion_com}
\end{figure*}

We evaluate amodal part completion from partial surface masks. Given the full object geometry and incomplete masks for all parts, OmniLatent predicts a complete full-object decomposition. The full geometry supplied to the completion adapter is generated by the base OmniLatent model rather than taken from ground truth. This setting is different from visible-surface segmentation: the model must preserve observed part evidence while inferring missing surface regions in a globally consistent way.

\paragraph{Evaluation protocol}
Following X-Part~\cite{yan2025x}, we evaluate on 200 samples from ObjaversePart-Tiny, each containing rendered images and ground-truth part geometries. All methods start from the full object geometry and partial surface masks, and produce completed full-object parts. Before metric computation, each object is normalized to $[-1,1]$. To make the evaluation robust to pose ambiguity, we rotate each prediction by $0^\circ$, $90^\circ$, $180^\circ$, and $270^\circ$, and report the best score among these orientations. We report Chamfer Distance (CD) and F-scores at distance thresholds 0.1 and 0.05. The scores are averaged over all test shapes.

We use two complementary views of the same outputs. The holistic metric evaluates the union of all completed parts and measures overall object-level geometric quality. The part-level metric evaluates each completed part against its corresponding ground-truth part geometry, then averages over parts and shapes. This separation is important because some baselines do not provide the same part-control interface as our input surface masks; for those methods, holistic quality remains comparable, while part-level evaluation is reported for controllable part-generation or part-decomposition methods whose outputs can be aligned with the provided 2D or 3D mask controls.

\paragraph{Baselines}
For holistic evaluation, we include Part123~\cite{liu2024part123}, HoloPart~\cite{yang2025holopart}, PartCrafter~\cite{lin2025partcrafter}, PartPacker~\cite{tang2025efficient}, OmniPart~\cite{yang2025omnipart}, and X-Part~\cite{yan2025x}. The part-level comparison uses controllable methods: SAMPart3D~\cite{yang2024sampart3d}, PartField~\cite{liu2025partfield}, HoloPart, OmniPart, and X-Part. SAMPart3D and PartField assign part labels on existing surfaces, while HoloPart, OmniPart, X-Part, and OmniLatent recover complete part structures. This comparison tests whether the method can both respect the provided surface evidence and complete missing part geometry.

\begin{table}[t]
\centering
\caption{\textbf{Holistic geometric evaluation for amodal part completion.} Given full geometry and partial surface masks, we evaluate the union of completed parts with Chamfer Distance (CD) and F-scores following X-Part.}
\label{tab:exp_completion_holistic}
% \small 
 \scalebox{0.9}{
\begin{tabular}{lccc}
\toprule
\textbf{Method} & CD$\downarrow$ & F-score-0.1$\uparrow$ & F-score-0.05$\uparrow$ \\
\midrule
% TripoSG  &  &  & 0.  \\
% Hunyuan2.1  &  &  & 0.  \\
Part123~\cite{liu2024part123}  & 0.42 & 0.36 & 0.20\\
HoloPart~\cite{yang2025holopart} & 0.09 & 0.88 & 0.73 \\
PartCrafter~\cite{lin2025partcrafter}  & 0.20 & 0.66 & 0.45 \\
PartPacker~\cite{tang2025efficient}  & 0.11 & 0.85 & 0.65 \\
OmniPart~\cite{yang2025omnipart}  & 0.08 & 0.91 & 0.77 \\
X-Part~\cite{yan2025x}      & 0.08 & 0.92 & 0.78 \\
\textbf{OmniLatent} (Ours) & \textbf{0.03} & \textbf{0.97} & \textbf{0.85} \\

\bottomrule
\end{tabular}
}
\end{table}

\begin{table}[t]
\centering
\caption{\textbf{Part-level evaluation for amodal part completion.} Given full geometry and partial surface masks, we evaluate completed parts against ground-truth part geometries with Chamfer Distance (CD) and F-scores following X-Part.}
\label{tab:exp_completion_part}
 \scalebox{0.93}{
% \small 
\begin{tabular}{lccc}
\toprule
\textbf{Method} & CD$\downarrow$ & F-score-0.1$\uparrow$ & F-score-0.05$\uparrow$ \\
\midrule
SAMPart3D~\cite{yang2024sampart3d}  & 0.15 & 0.73 & 0.63 \\
PartField~\cite{liu2025partfield}  & 0.17 & 0.68 & 0.57 \\
HoloPart~\cite{yang2025holopart}  & 0.26 & 0.59 & 0.43  \\
OmniPart~\cite{yang2025omnipart}  & 0.23 & 0.63 & 0.46 \\
X-Part~\cite{yan2025x}     & 0.11 & 0.80 & 0.71  \\
Ours (w/o surface RoPE) & 0.15 & 0.75 & 0.64  \\
Ours (w/o Full-attn) & 0.13 & 0.77 & 0.68  \\
\textbf{OmniLatent} (Ours) & \textbf{0.08} & \textbf{0.85} & \textbf{0.78}  \\

\bottomrule
\end{tabular}
}
\end{table}

\paragraph{Results}
\Cref{tab:exp_completion_holistic} shows that OmniLatent achieves the best holistic geometric quality, reducing CD to 0.03 and improving F-score at both thresholds. \Cref{tab:exp_completion_part} further shows that OmniLatent achieves the best part-level completion quality. Compared with X-Part, OmniLatent reduces CD from 0.11 to 0.08 and improves F-score-0.05 from 0.71 to 0.78. The two ablations confirm the importance of the completion-specific design. Removing surface RoPE weakens the anchoring of visible mask evidence, increasing CD from 0.08 to 0.15 and reducing F-score-0.05 from 0.78 to 0.64. Removing full attention limits cross-part communication, increasing CD to 0.13 and reducing F-score-0.05 to 0.68. Qualitatively, \Cref{fig:part_completion_com} shows that OmniLatent preserves visible part evidence while completing missing regions into coherent full-object parts.

The completion model benefits from three design choices. Dynamic token allocation gives larger visible parts more capacity. No-RoPE completion tokens allow the model to infer unobserved part regions without forcing them to occupy preassigned visible positions. Alternating part-local and global attention lets the model complete fine details within each part while maintaining consistency across neighboring parts and the overall shape.

\subsection{Supporting Generation Quality}
\label{sec:4-4}

\begin{figure}
\centering
\includegraphics[width=\linewidth]{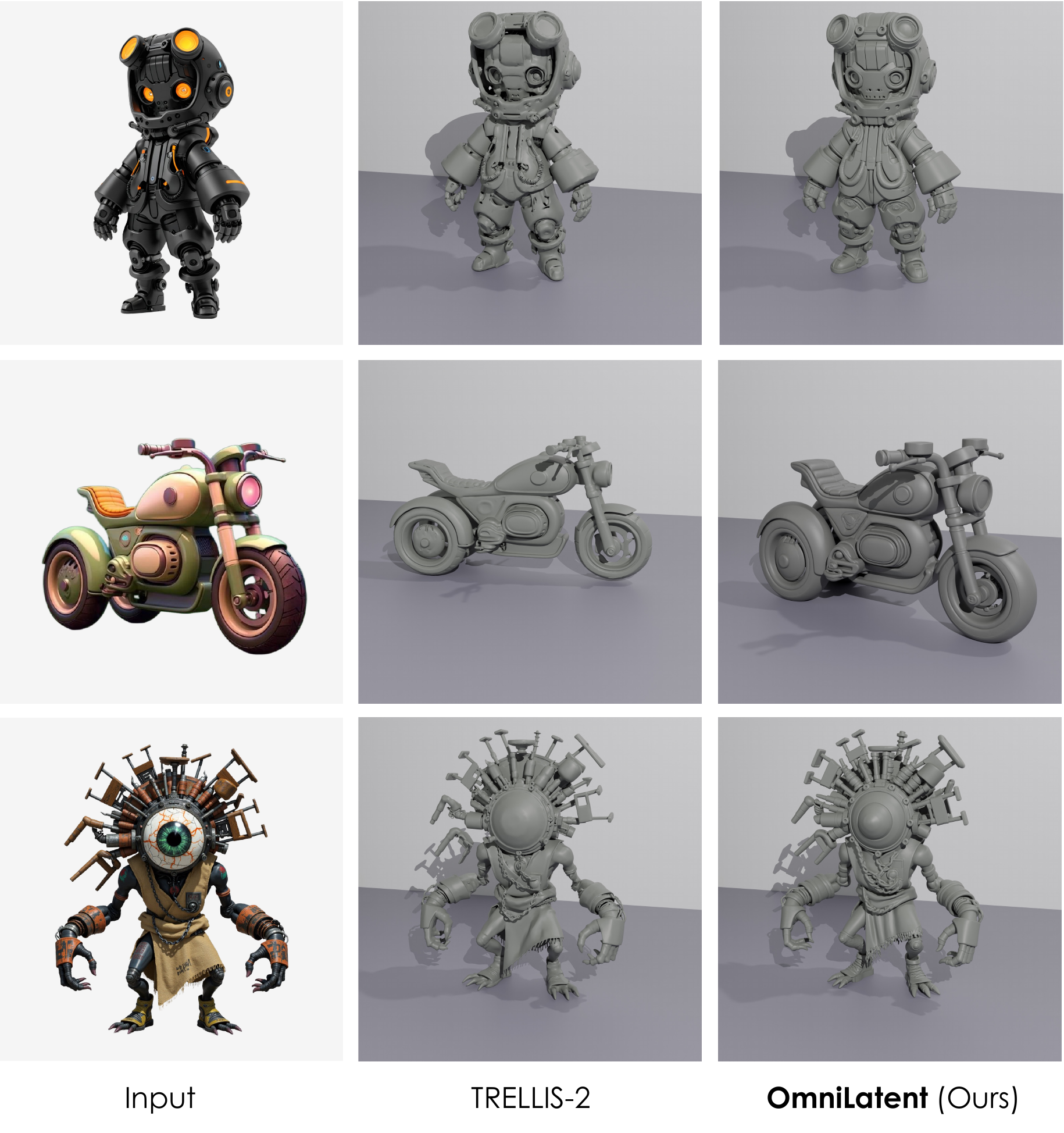}
\caption{\textbf{Visual comparison of geometry generation}. Despite its unified design, our method still produces high-quality generations, providing strong geometric understanding that benefits part decomposition.}
\label{fig:visual_geometry}
\end{figure}

\begin{figure*}
\centering
\includegraphics[width=\linewidth]{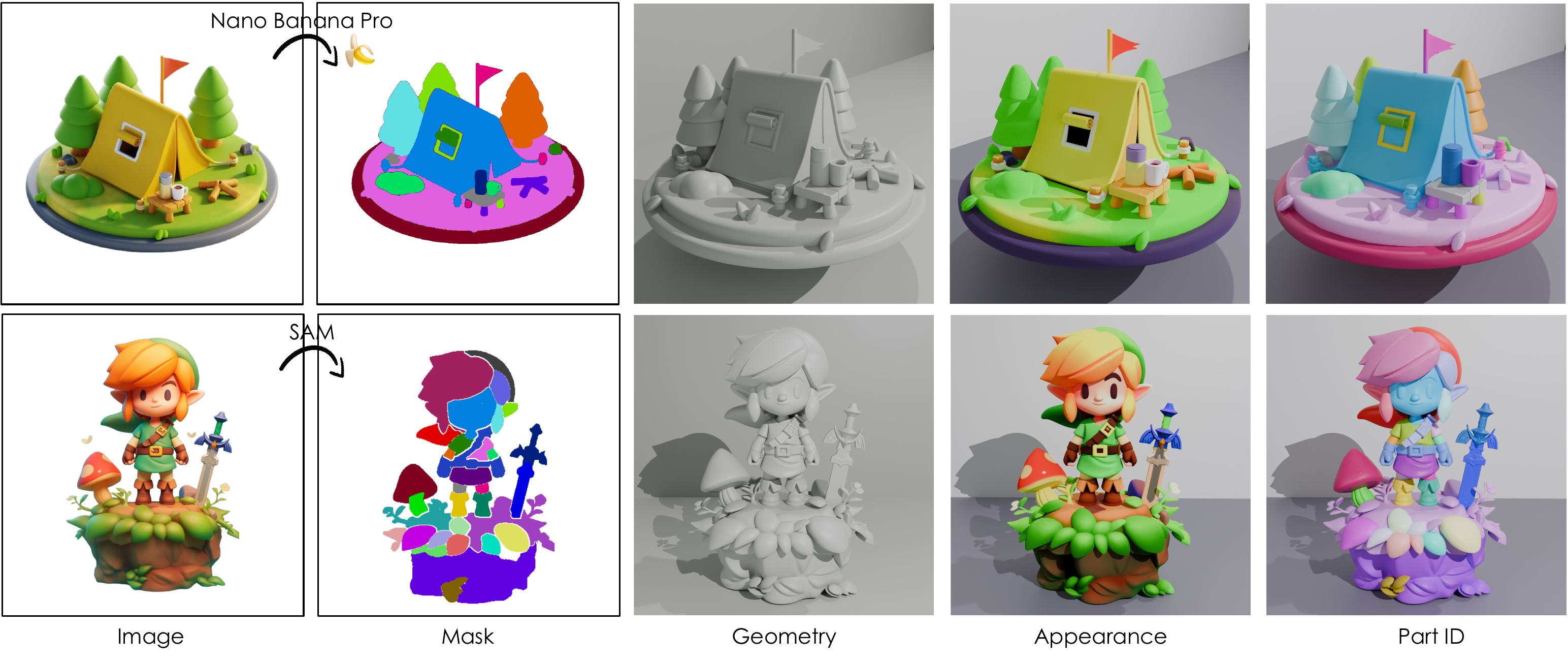}
\caption{\textbf{Unified generation and mask-guided surface part segmentation.} OmniLatent jointly generates geometry, deformation, appearance, and surface part IDs from a single image, and can use 2D segmentation masks as controllable guidance for part granularity.}
\label{fig:exp_part_tex_geo}
\end{figure*}

Although the main goal of OmniLatent is part decomposition, geometry and appearance quality remain important because they provide the structural and visual context used by the part decoder. As shown in \Cref{fig:visual_geometry,fig:exp_part_tex_geo}, the same generated OmniLatent sample provides geometry, appearance, and surface part IDs, while optional 2D masks can control the desired part granularity. We therefore report supporting evaluations for geometry and appearance generation, and analyze test-time token scaling inherited from the voxel-query representation.

\paragraph{Geometry}
We compare against Michelangelo~\cite{zhao2024michelangelo}, Craftman 1.5~\cite{li2024craftsman}, TRELLIS~\cite{xiang2024structured}, TRELLIS-2~\cite{xiang2025native}, Hunyuan3D-2~\cite{zhao2025hunyuan3d}, Hi3DGen~\cite{ye2025hi3dgen}, and Direct3D-s2~\cite{wu2025direct3d}. These baselines cover representative vecset-based, voxel-based, and image-conditioned 3D generation paradigms. Following the evaluation protocol in Hunyuan3D-2, we construct a benchmark of 450 challenging and highly detailed assets and use ULIP~\cite{xue2023ulip} and Uni3D~\cite{zhou2023uni3d} to measure text-mesh and image-mesh similarities. The results in \Cref{tab:geo_generation} show that OmniLatent maintains competitive geometry generation while jointly modeling part IDs. As shown in \Cref{fig:visual_geometry}, the generated surfaces preserve coherent global structure and local details.

\paragraph{Appearance}
We evaluate image-conditioned appearance generation on the MaterialMVP test set, which contains 268 images with ground-truth geometries. We compare with Paint3D~\cite{zeng2024paint3d}, TexGen~\cite{yu2024texgen}, Hunyuan3D-2~\cite{zhao2025hunyuan3d}, RomanTex~\cite{feng2025romantex}, and MaterialMVP~\cite{he2025materialmvp}. Since these baselines perform texturing on provided geometries, we use the existing geometries as the source for our voxel queries to maintain a consistent evaluation setting. We report c-FID, LPIPS, CMMD, and CLIP-I in \Cref{tab:texture_gen}. The results indicate that the OmniLatent space preserves appearance fidelity while providing aligned visual cues for part decomposition.

\begin{table}
\centering
\small
\caption{Quantitative comparison of geometry generation.}
\setlength{\tabcolsep}{5pt}
\resizebox{1.0\linewidth}{!}{
\begin{tabular}{c|cccc}
\hline
               \textbf{Method}         & \textbf{ULIP-T}  & \textbf{ULIP-I} & \textbf{Uni-T} & \textbf{Uni-I}\\ \hline
Michelangelo~\citep{zhao2024michelangelo} & 0.075 & 0.115 & 0.213 & 0.261 \\
Craftsman 1.5~\citep{li2024craftsman}    & 0.074 & 0.129 & 0.237 & 0.298 \\
TRELLIS~\citep{xiang2024structured}      & 0.076 & 0.126 & 0.249 & 0.311 \\
Hunyuan3D 2.0~\citep{zhao2025hunyuan3d} & \textbf{0.077} & \textbf{0.130} & {0.251} & {0.315} \\ 
Hi3DGen~\cite{ye2025hi3dgen}  & 0.066 & 0.112  & 0.246 & 0.299 \\
Direct3D-s2~\cite{wu2025direct3d} & 0.074 & 0.122 & 0.247 & 0.314  \\
% \hline
% Tripo 2.5                & 0.071 & \textbf{0.130} & 0.246  \\
% Rodin 1.5                & 0.076 & 0.124  & \underline{0.251}  \\
% Meshy 5                  & \textbf{0.071} & \underline{0.130} & \textbf{0.259} \\ 
\hline
% LATTICE-0.6B                  & \textbf{0.071} & \underline{0.133} & \textbf{0.259}  \\
\textbf{OmniLatent} (Ours)                  & 0.073 & 0.126  & \textbf{0.252}   & \textbf{0.318}  \\ 
% LATTICE-4.5B                  & \textbf{0.071} & \underline{0.133} & \textbf{0.259}  \\
\hline
\end{tabular}
}
% \caption{Numerical comparison of geometry generation performance on LATTICE-bench(G) against various aspects.}

% uni-T 和 uni-I 是第三次爬坡
% ulip-T 是第三次爬坡，ulip-I 是第二次爬坡

\label{tab:geo_generation}
\end{table}

\begin{table}[t]
    \centering
    \caption{Quantitative comparison of appearance generation.}
    \setlength{\tabcolsep}{5pt}\small
    \resizebox{1.0\linewidth}{!}{
    \begin{tabular}{c|cccccc}
    \toprule 
    \textbf{Method}     &  \textbf{cFID$\downarrow$} & \textbf{CMMD$\downarrow$} & \textbf{CLIP$\uparrow$} & \textbf{LPIPS$\downarrow$}  \\ \midrule
    Paint3D \cite{zeng2024paint3d}     & 26.86      & 2.400 & 0.887      & 0.126   \\
    TexGen \cite{yu2024texgen}      & 28.23      & 2.447 & 0.882 & 0.133     \\ 
    Hunyuan3D-2 \cite{zhao2025hunyuan3d}     & 26.43     & 2.318 & 0.889 & 0.126       \\ 
    RomanTex \cite{feng2025romantex}     & 24.78      & 2.191 & 0.891 & 0.121       \\ 
    MaterialMVP \cite{he2025materialmvp}      & {24.78}      & {2.191} & \textbf{0.921} & {0.121}       \\ 
    \hline
    \textbf{OmniLatent} (Ours)       & \textbf{22.95}      & \textbf{2.079} & {0.903} & \textbf{0.111}       \\ \bottomrule
    \end{tabular}
    }

    \label{tab:texture_gen}
\end{table}

\paragraph{Test-time token scaling}
We evaluate the behavior of OmniLatent across different inference token counts. The generated geometry becomes progressively more detailed as the number of inference tokens increases from 3,072 to 20,480, even though the VAE and DiT are trained with maximum sequence lengths of 6,144 and 12,248 tokens, respectively. This extrapolation behavior comes from the ordered voxel-query representation and random token sampling during training, which allow the model to learn both global structure and local details under varying token densities.

\subsection{Ablation Analysis}
\label{sec:4-5}

\begin{figure}
\centering
\includegraphics[width=\linewidth]{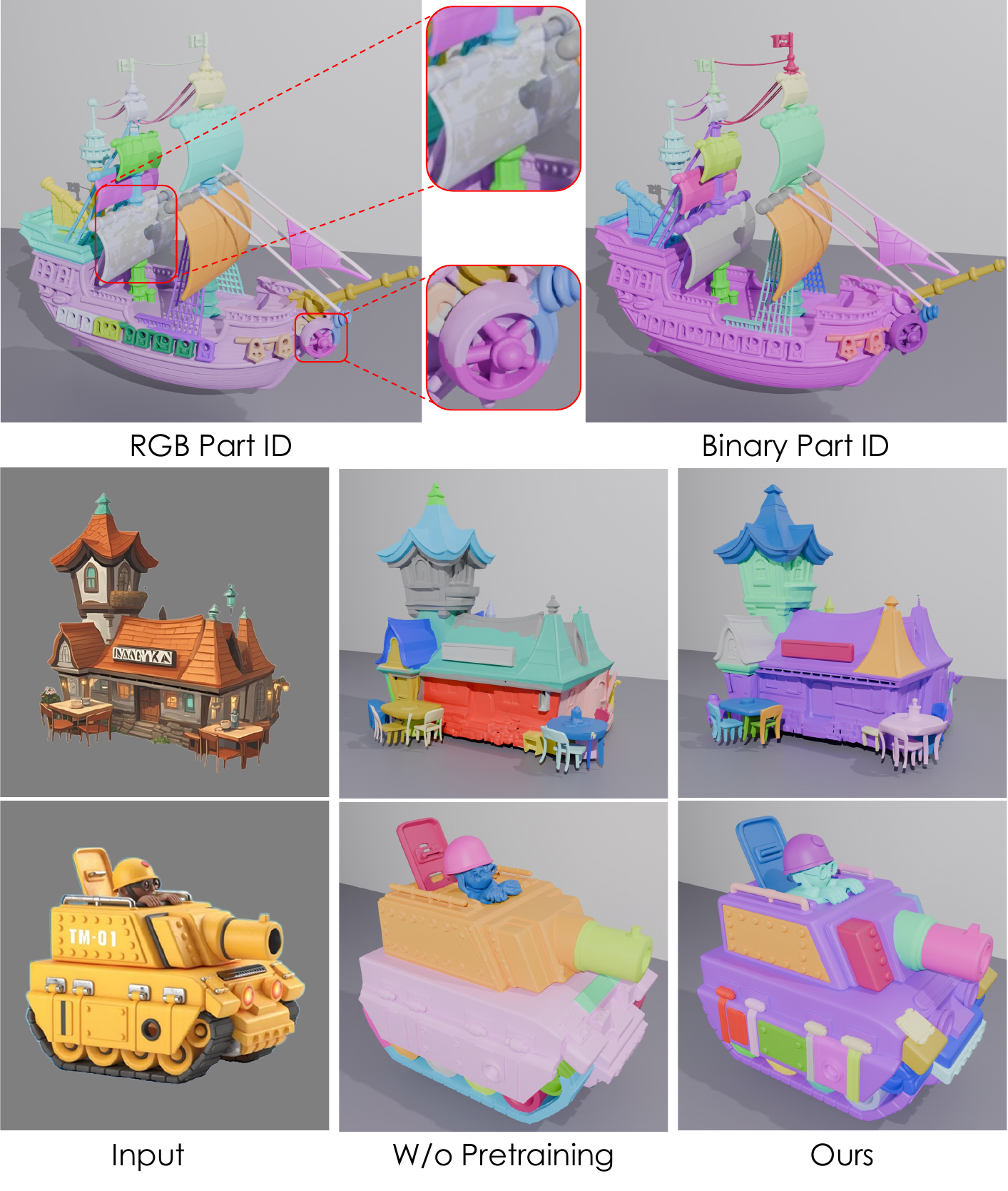}
\caption{Ablation study on Part ID representation and pretraining.}
\label{fig:part_abl}
\end{figure}

\paragraph{Binary part ID}
We adopt a 6-bit binary representation for part IDs. As illustrated in \Cref{fig:part_abl}, binary IDs produce cleaner and more stable part assignments than RGB-coded masks. RGB encodings can introduce small color deviations during denoising, causing a single part to fragment into multiple inconsistent IDs.

\begin{figure}
\centering
\includegraphics[width=\linewidth]{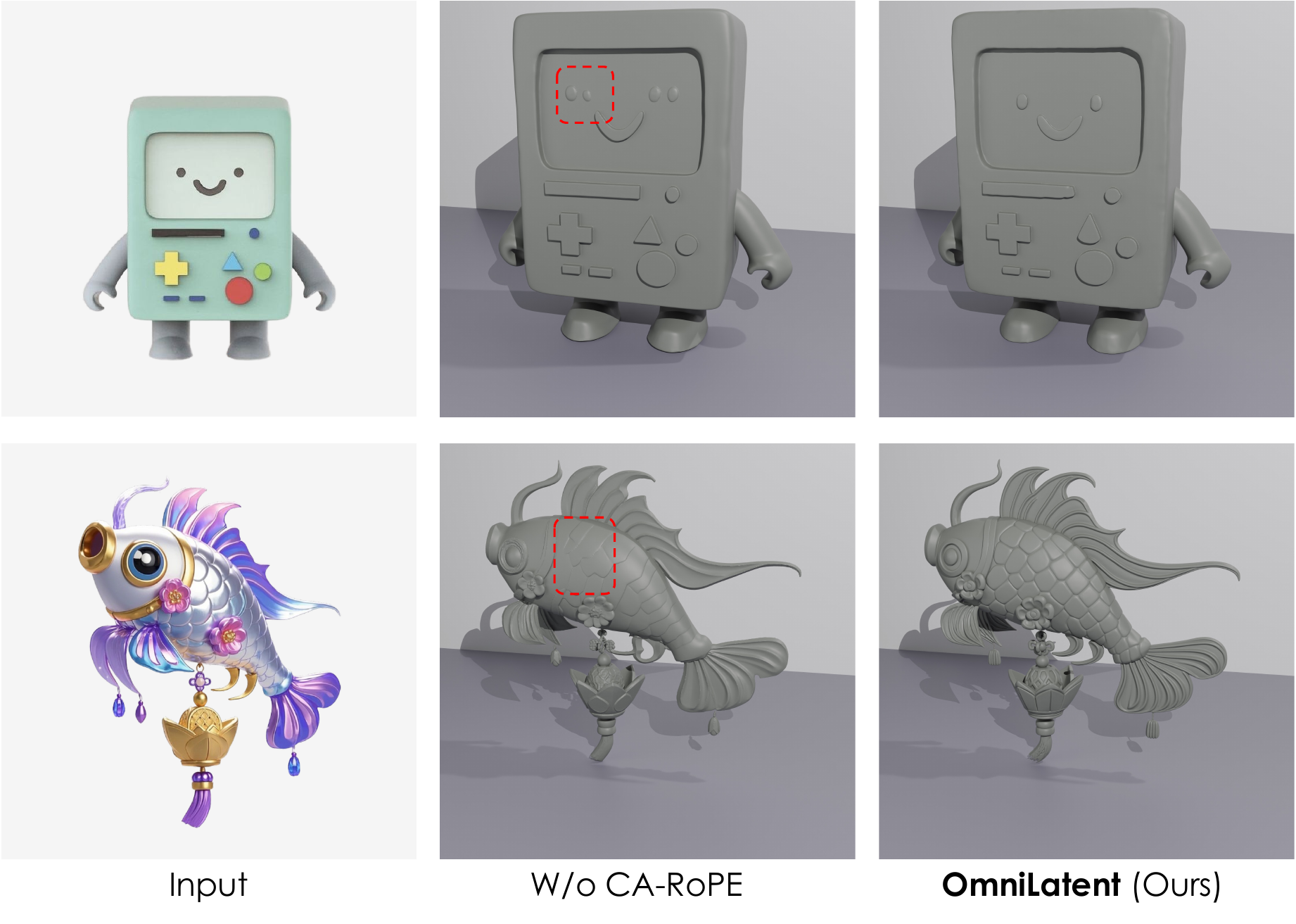}
\caption{Ablation study on Condition-Aware 3D Rotary Position Embedding.}
\label{fig:exp_geo_abl}
\end{figure}

\paragraph{Geometry and appearance pretraining}
We evaluate the importance of the progressive training strategy by comparing the full model with variants trained without geometry and appearance pretraining. As shown in \Cref{fig:part_abl}, removing these priors leads to noisier part regions and less stable boundaries. This supports our central design: geometry and appearance are not separate headline tasks, but provide the latent context that makes part decomposition more reliable.

\paragraph{Condition-Aware RoPE}
We ablate CA-RoPE in \Cref{tab:ablation_part} and \Cref{fig:exp_geo_abl}. Without condition-aware positional indexing, image and segmentation mask conditions are less consistently aligned with the 3D latent tokens, which weakens the transfer of spatial cues to generated geometry and part boundaries. Removing CA-RoPE reduces the average mIoU from 74.18 to 69.20, confirming its importance for both mask-guided segmentation and mask-free single-image generation.

\paragraph{Unified latent modeling}
To isolate the benefit of the shared latent space, we compare OmniLatent with a Part ID-only Latent baseline that learns part IDs without the unified geometry and appearance latent priors. \Cref{tab:ablation_part} shows a clear improvement from 67.69 to 74.18 average mIoU, indicating that the data-rich geometry and appearance modalities provide useful structural and visual priors for data-scarce part decomposition. \Cref{fig:exp_geo_abl} also visualizes how joint geometry-appearance latent modeling improves the coherence between synthesized texture and generated structure.

% \begin{table*}[t]
% \centering
% \caption{Ablation study on unified latent modeling and CA-RoPE for surface part segmentation on PartObjaverse-Tiny-wt, reported in mIoU (\%).}
% \resizebox{0.95\linewidth}{!}{
% \begin{tabular}{c|c|cccccccc}
% \hline
%  \textbf{Method}  & \textbf{Avg.} $\uparrow$ &  Human-Shape& Animals& Daily-Used& Buildings & Transportations & Plants& Food & Electronics         \\  \hline
%  Part ID-only Latent & 67.69 & 67.34 & 76.34 & 68.89 & 51.34 & 59.27 & 74.59 & 78.84 & 64.90 \\
%  W/o CA-RoPE & 69.20 & 74.99 & 72.97 & 72.60 & 52.33 & 60.49 & 74.94 & 80.12 & 65.19 \\
%  \textbf{OmniLatent} (Ours) & \textbf{74.18} & \textbf{79.36} & \textbf{82.68} & \textbf{76.98} & \textbf{58.94} & \textbf{65.04} & \textbf{78.73} & \textbf{80.72} & \textbf{70.98} \\ \hline
% \end{tabular}
% }
% \label{tab:ablation_part}
% \end{table*}

\begin{table}[t]
\centering
\caption{Ablation study on unified latent modeling and CA-RoPE for surface part segmentation on PartObjaverse-Tiny-wt, reported in mIoU (\%).}
\resizebox{1\linewidth}{!}{
\begin{tabular}{c|ccc}
\hline
\textbf{Category} & \textbf{Part ID-only Latent} & \textbf{W/o CA-RoPE} & \textbf{OmniLatent (Ours)} \\ \hline
Avg.              & 67.69 & 69.20 & \textbf{74.18} \\
Human-Shape       & 67.34 & 74.99 & \textbf{79.36} \\
Animals           & 76.34 & 72.97 & \textbf{82.68} \\
Daily-Used        & 68.89 & 72.60 & \textbf{76.98} \\
Buildings         & 51.34 & 52.33 & \textbf{58.94} \\
Transportations   & 59.27 & 60.49 & \textbf{65.04} \\
Plants            & 74.59 & 74.94 & \textbf{78.73} \\
Food              & 78.84 & 80.12 & \textbf{80.72} \\
Electronics       & 64.90 & 65.19 & \textbf{70.98} \\ \hline
\end{tabular}
}
\label{tab:ablation_part}
\end{table}

\subsection{Computational Complexity}
OmniLatent generates geometry, deformation, appearance, and part IDs from a single image in approximately 30 seconds with 8192 latent tokens. For comparison, a representative separate pipeline requires approximately 30 seconds for geometry generation, 30 seconds for texture generation, and 20 seconds for Part-ID segmentation, totaling about 80 seconds. This is substantially more efficient than running separate geometry, texture, and post-hoc part segmentation stages, and it avoids cross-stage error accumulation by decoding all attributes from the same latent sample.

\section{Conclusion and Limitations}
We introduced OmniLatent, a unified latent modeling framework for 3D part decomposition. By encoding geometry, deformation, appearance, and binary part IDs into the same OmniLatent space, OmniLatent transfers geometric and visual priors from data-rich modalities to data-scarce part learning. The resulting base model generates geometry, deformation, texture, and surface part IDs from a single image in one pass, supports optional mask guidance, and can be tuned for point-prompted full segmentation, segmentation refinement, part subdivision, and amodal part completion.

\begin{figure}[t]
\centering
% Add a third panel for the 2D-mask-induced back-side error when the result asset is available.
\includegraphics[width=\linewidth]{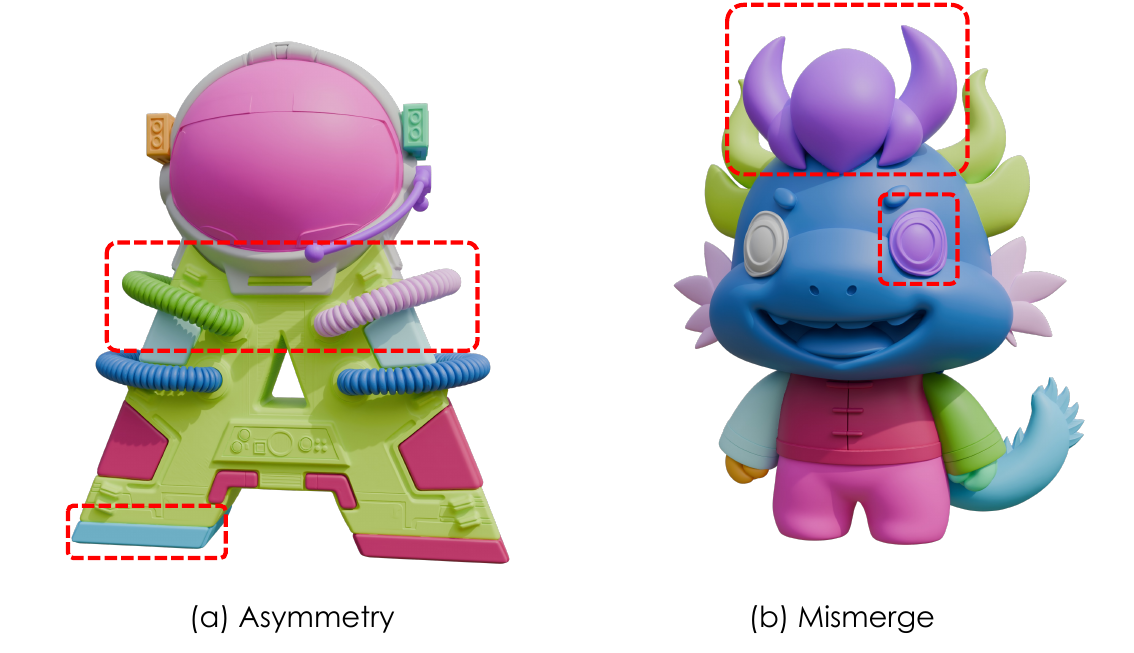}
\caption{\textbf{Limitations.} OmniLatent can still produce asymmetric part layouts on symmetric structures and may mistakenly merge semantically distinct regions into a single part.}
\label{fig:limitations}
\end{figure}

OmniLatent also exposes several directions for future work. The quality of autonomous segmentation still depends on the diversity and semantic consistency of available part-level 3D data. As shown in \Cref{fig:limitations}, one limitation is symmetry consistency: for objects with repeated or bilateral structures, the model may occasionally assign different part layouts to geometrically corresponding regions. Another limitation is erroneous merging, where two components that share similar geometry, appearance, or spatial context may be grouped into the same part even though they should remain separate for editing. A third limitation arises from view-dependent mask guidance: a 2D mask can induce back-side errors by splitting surface regions that should remain connected in 3D. These cases suggest that future work could incorporate stronger symmetry-aware priors, connectivity constraints, or interactive correction mechanisms. Extending the same unified latent formulation to additional production attributes, such as materials, UVs, or articulation, is another promising direction for building more editable 3D assets.

\bibliographystyle{ACM-Reference-Format}
\bibliography{sample-bibliography}

@String{Computing = "Computing" }

@String{Computer = "{IEEE} Computer" }

@article{zhang2024clay,
  title={CLAY: A Controllable Large-scale Generative Model for Creating High-quality 3D Assets},
  author={Zhang, Longwen and Wang, Ziyu and Zhang, Qixuan and Qiu, Qiwei and Pang, Anqi and Jiang, Haoran and Yang, Wei and Xu, Lan and Yu, Jingyi},
  journal={ACM Transactions on Graphics (TOG)},
  volume={43},
  number={4},
  pages={1--20},
  year={2024},
}

@article{li2024craftsman,
  title={CraftsMan: High-fidelity Mesh Generation with 3D Native Generation and Interactive Geometry Refiner},
  author={Li, Weiyu and Liu, Jiarui and Chen, Rui and Liang, Yixun and Chen, Xuelin and Tan, Ping and Long, Xiaoxiao},
  journal={arXiv preprint arXiv:2405.14979},
  year={2024}
}

@inproceedings{poole2022dreamfusion,
  title={DreamFusion: Text-to-3D using 2D Diffusion},
  author={Poole, Ben and Jain, Ajay and Barron, Jonathan T and Mildenhall, Ben},
  booktitle={ICLR},
  year={2023}
}

@article{zhang20233dshape2vecset,
  title={3dshape2vecset: A 3d shape representation for neural fields and generative diffusion models},
  author={Zhang, Biao and Tang, Jiapeng and Niessner, Matthias and Wonka, Peter},
  journal={ACM Transactions On Graphics (TOG)},
  volume={42},
  number={4},
  pages={1--16},
  year={2023},
}

@inproceedings{long2024wonder3d,
  title={Wonder3d: Single image to 3d using cross-domain diffusion},
  author={Long, Xiaoxiao and Guo, Yuan-Chen and Lin, Cheng and Liu, Yuan and Dou, Zhiyang and Liu, Lingjie and Ma, Yuexin and Zhang, Song-Hai and Habermann, Marc and Theobalt, Christian and others},
  booktitle={CVPR},
  year={2024}
}

@inproceedings{wu2024direct3d,
  title={Direct3D: Scalable Image-to-3D Generation via 3D Latent Diffusion Transformer},
  author={Wu, Shuang and Lin, Youtian and Zhang, Feihu and Zeng, Yifei and Xu, Jingxi and Torr, Philip and Cao, Xun and Yao, Yao},
  booktitle={NeurIPS},
  year={2024}
}

@inproceedings{zhao2024michelangelo,
  title={Michelangelo: Conditional 3d shape generation based on shape-image-text aligned latent representation},
  author={Zhao, Zibo and Liu, Wen and Chen, Xin and Zeng, Xianfang and Wang, Rui and Cheng, Pei and Fu, Bin and Chen, Tao and Yu, Gang and Gao, Shenghua},
  booktitle={NeurIPS},
  year={2024}
}

@inproceedings{dai2017shape,
  title={Shape completion using 3d-encoder-predictor cnns and shape synthesis},
  author={Dai, Angela and Ruizhongtai Qi, Charles and Nie{\ss}ner, Matthias},
  booktitle={CVPR},
  year={2017}
}

@inproceedings{lan2025ln3diff,
  title={Ln3diff: Scalable latent neural fields diffusion for speedy 3d generation},
  author={Lan, Yushi and Hong, Fangzhou and Yang, Shuai and Zhou, Shangchen and Meng, Xuyi and Dai, Bo and Pan, Xingang and Loy, Chen Change},
  booktitle={ECCV},
  year={2025},
}

@inproceedings{zhang2025compress3d,
  title={Compress3D: a compressed latent space for 3D generation from a single image},
  author={Zhang, Bowen and Yang, Tianyu and Li, Yu and Zhang, Lei and Zhao, Xi},
  booktitle={ECCV},
  year={2025},
}

@inproceedings{huang2025midi,
  title={Midi: Multi-instance diffusion for single image to 3d scene generation},
  author={Huang, Zehuan and Guo, Yuan-Chen and An, Xingqiao and Yang, Yunhan and Li, Yangguang and Zou, Zi-Xin and Liang, Ding and Liu, Xihui and Cao, Yan-Pei and Sheng, Lu},
  booktitle={CVPR},
  year={2025}
}

@incollection{lorensen1998marching,
  title={Marching cubes: A high resolution 3D surface construction algorithm},
  author={Lorensen, William E and Cline, Harvey E},
  booktitle={Seminal graphics: pioneering efforts that shaped the field},
  year={1998}
}

@inproceedings{deitke2023objaverse,
  title={Objaverse: A universe of annotated 3d objects},
  author={Deitke, Matt and Schwenk, Dustin and Salvador, Jordi and Weihs, Luca and Michel, Oscar and VanderBilt, Eli and Schmidt, Ludwig and Ehsani, Kiana and Kembhavi, Aniruddha and Farhadi, Ali},
  booktitle={CVPR},
  year={2023}
}

@article{yang2024sampart3d,
  title={Sampart3d: Segment any part in 3d objects},
  author={Yang, Yunhan and Huang, Yukun and Guo, Yuan-Chen and Lu, Liangjun and Wu, Xiaoyang and Lam, Edmund Y and Cao, Yan-Pei and Liu, Xihui},
  journal={arXiv preprint arXiv:2411.07184},
  year={2024}
}

@inproceedings{liu2024part123,
  title={Part123: part-aware 3d reconstruction from a single-view image},
  author={Liu, Anran and Lin, Cheng and Liu, Yuan and Long, Xiaoxiao and Dou, Zhiyang and Guo, Hao-Xiang and Luo, Ping and Wang, Wenping},
  booktitle={ACM SIGGRAPH},
  year={2024}
}

@article{chen2024partgen,
  title={PartGen: Part-level 3D Generation and Reconstruction with Multi-View Diffusion Models},
  author={Chen, Minghao and Shapovalov, Roman and Laina, Iro and Monnier, Tom and Wang, Jianyuan and Novotny, David and Vedaldi, Andrea},
  journal={arXiv preprint arXiv:2412.18608},
  year={2024}
}

@inproceedings{hui2022neural,
  title={Neural wavelet-domain diffusion for 3d shape generation},
  author={Hui, Ka-Hei and Li, Ruihui and Hu, Jingyu and Fu, Chi-Wing},
  booktitle={SIGGRAPH Asia},
  year={2022}
}

@inproceedings{koo2023salad,
  title={Salad: Part-level latent diffusion for 3d shape generation and manipulation},
  author={Koo, Juil and Yoo, Seungwoo and Nguyen, Minh Hieu and Sung, Minhyuk},
  booktitle={ICCV},
  year={2023}
}

@inproceedings{chou2023diffusion,
  title={Diffusion-sdf: Conditional generative modeling of signed distance functions},
  author={Chou, Gene and Bahat, Yuval and Heide, Felix},
  booktitle={ICCV},
  year={2023}
}

@inproceedings{shim2023diffusion,
  title={Diffusion-based signed distance fields for 3d shape generation},
  author={Shim, Jaehyeok and Kang, Changwoo and Joo, Kyungdon},
  booktitle={CVPR},
  year={2023}
}

@inproceedings{li2023diffusion,
  title={Diffusion-sdf: Text-to-shape via voxelized diffusion},
  author={Li, Muheng and Duan, Yueqi and Zhou, Jie and Lu, Jiwen},
  booktitle={CVPR},
  year={2023}
}

@article{li2025triposg,
  title={TripoSG: High-Fidelity 3D Shape Synthesis using Large-Scale Rectified Flow Models},
  author={Li, Yangguang and Zou, Zi-Xin and Liu, Zexiang and Wang, Dehu and Liang, Yuan and Yu, Zhipeng and Liu, Xingchao and Guo, Yuan-Chen and Liang, Ding and Ouyang, Wanli and others},
  journal={arXiv preprint arXiv:2502.06608},
  year={2025}
}

@article{xiang2024structured,
  title={Structured 3d latents for scalable and versatile 3d generation},
  author={Xiang, Jianfeng and Lv, Zelong and Xu, Sicheng and Deng, Yu and Wang, Ruicheng and Zhang, Bowen and Chen, Dong and Tong, Xin and Yang, Jiaolong},
  journal={arXiv preprint arXiv:2412.01506},
  year={2024},
}

@inproceedings{rombach2022high,
  title={High-resolution image synthesis with latent diffusion models},
  author={Rombach, Robin and Blattmann, Andreas and Lorenz, Dominik and Esser, Patrick and Ommer, Bj{\"o}rn},
  booktitle={CVPR},
  year={2022}
}

@inproceedings{liu2023zero1to3,
  title={Zero-1-to-3: Zero-shot one image to 3d object},
  author={Liu, Ruoshi and Wu, Rundi and Van Hoorick, Basile and Tokmakov, Pavel and Zakharov, Sergey and Vondrick, Carl},
  booktitle={ICCV},
  year={2023}
}

@article{shi2023zero123plus,
  title={Zero123++: a single image to consistent multi-view diffusion base model},
  author={Shi, Ruoxi and Chen, Hansheng and Zhang, Zhuoyang and Liu, Minghua and Xu, Chao and Wei, Xinyue and Chen, Linghao and Zeng, Chong and Su, Hao},
  journal={arXiv preprint arXiv:2310.15110},
  year={2023}
}

@inproceedings{liu2023syncdreamer,
  title={SyncDreamer: Learning to Generate Multiview-consistent Images from a Single-view Image},
  author={Liu, Yuan and Lin, Cheng and Zeng, Zijiao and Long, Xiaoxiao and Liu, Lingjie and Komura, Taku and Wang, Wenping},
  booktitle={ICLR},
  year={2024}
}

@inproceedings{wang2023prolificdreamer,
  title={Prolificdreamer: High-fidelity and diverse text-to-3d generation with variational score distillation},
  author={Wang, Zhengyi and Lu, Cheng and Wang, Yikai and Bao, Fan and Li, Chongxuan and Su, Hang and Zhu, Jun},
  booktitle={NeurIPS},
  year={2023}
}

@inproceedings{lin2023magic3d,
  title={Magic3d: High-resolution text-to-3d content creation},
  author={Lin, Chen-Hsuan and Gao, Jun and Tang, Luming and Takikawa, Towaki and Zeng, Xiaohui and Huang, Xun and Kreis, Karsten and Fidler, Sanja and Liu, Ming-Yu and Lin, Tsung-Yi},
  booktitle={CVPR},
  year={2023}
}

@inproceedings{chen2023fantasia3d,
  title={Fantasia3d: Disentangling geometry and appearance for high-quality text-to-3d content creation},
  author={Chen, Rui and Chen, Yongwei and Jiao, Ningxin and Jia, Kui},
  booktitle={ICCV},
  year={2023}
}

@inproceedings{qian2023magic123,
  title={Magic123: One image to high-quality 3d object generation using both 2d and 3d diffusion priors},
  author={Qian, Guocheng and Mai, Jinjie and Hamdi, Abdullah and Ren, Jian and Siarohin, Aliaksandr and Li, Bing and Lee, Hsin-Ying and Skorokhodov, Ivan and Wonka, Peter and Tulyakov, Sergey and others},
  booktitle={ICLR},
  year={2024}
}

@inproceedings{deitke2022objaverse,
  title={Objaverse: A universe of annotated 3d objects},
  author={Deitke, Matt and Schwenk, Dustin and Salvador, Jordi and Weihs, Luca and Michel, Oscar and VanderBilt, Eli and Schmidt, Ludwig and Ehsani, Kiana and Kembhavi, Aniruddha and Farhadi, Ali},
  booktitle={CVPR},
  year={2023}
}

@inproceedings{liu2023one2345,
  title={One-2-3-45++: Fast Single Image to 3D Objects with Consistent Multi-View Generation and 3D Diffusion}, 
  author={Minghua Liu and Ruoxi Shi and Linghao Chen and Zhuoyang Zhang and Chao Xu and Xinyue Wei and Hansheng Chen and Chong Zeng and Jiayuan Gu and Hao Su},
  booktitle={CVPR},
  year={2024},
}

@inproceedings{chen2024meshanything,
  title={Meshanything: Artist-created mesh generation with autoregressive transformers},
  author={Chen, Yiwen and He, Tong and Huang, Di and Ye, Weicai and Chen, Sijin and Tang, Jiaxiang and Chen, Xin and Cai, Zhongang and Yang, Lei and Yu, Gang and others},
  booktitle={ICLR},
  year={2025},
}

@article{chen2024meshanythingv2,
  title={Meshanything v2: Artist-created mesh generation with adjacent mesh tokenization},
  author={Chen, Yiwen and Wang, Yikai and Luo, Yihao and Wang, Zhengyi and Chen, Zilong and Zhu, Jun and Zhang, Chi and Lin, Guosheng},
  journal={arXiv preprint arXiv:2408.02555},
  year={2024}
}

@inproceedings{tang2024edgerunner,
  title={Edgerunner: Auto-regressive auto-encoder for artistic mesh generation},
  author={Tang, Jiaxiang and Li, Zhaoshuo and Hao, Zekun and Liu, Xian and Zeng, Gang and Liu, Ming-Yu and Zhang, Qinsheng},
  booktitle={ICLR},
  year={2025},
}

@inproceedings{chen2024meshxl,
  title={Meshxl: Neural coordinate field for generative 3d foundation models},
  author={Chen, Sijin and Chen, Xin and Pang, Anqi and Zeng, Xianfang and Cheng, Wei and Fu, Yijun and Yin, Fukun and Wang, Billzb and Yu, Jingyi and Yu, Gang and others},
  booktitle={NeurIPS},
  year={2024}
}

@article{weng2024scaling,
  title={Scaling mesh generation via compressive tokenization},
  author={Weng, Haohan and Zhao, Zibo and Lei, Biwen and Yang, Xianghui and Liu, Jian and Lai, Zeqiang and Chen, Zhuo and Liu, Yuhong and Jiang, Jie and Guo, Chunchao and others},
  journal={arXiv preprint arXiv:2411.07025},
  year={2024}
}

@article{hao2024meshtron,
  title={Meshtron: High-Fidelity, Artist-Like 3D Mesh Generation at Scale},
  author={Hao, Zekun and Romero, David W and Lin, Tsung-Yi and Liu, Ming-Yu},
  journal={arXiv preprint arXiv:2412.09548},
  year={2024}
}

@inproceedings{siddiqui2024meshgpt,
  title={Meshgpt: Generating triangle meshes with decoder-only transformers},
  author={Siddiqui, Yawar and Alliegro, Antonio and Artemov, Alexey and Tommasi, Tatiana and Sirigatti, Daniele and Rosov, Vladislav and Dai, Angela and Nie{\ss}ner, Matthias},
  booktitle={CVPR},
  year={2024}
}

@inproceedings{zou2024triplane,
  title={Triplane meets gaussian splatting: Fast and generalizable single-view 3d reconstruction with transformers},
  author={Zou, Zi-Xin and Yu, Zhipeng and Guo, Yuan-Chen and Li, Yangguang and Liang, Ding and Cao, Yan-Pei and Zhang, Song-Hai},
  booktitle={CVPR},
  year={2024}
}

@article{yang2025holopart,
  title={HoloPart: Generative 3D Part Amodal Segmentation},
  author={Yang, Yunhan and Guo, Yuan-Chen and Huang, Yukun and Zou, Zi-Xin and Yu, Zhipeng and Li, Yangguang and Cao, Yan-Pei and Liu, Xihui},
  journal={arXiv preprint arXiv:2504.07943},
  year={2025}
}

@article{liu2025partfield,
  title={PARTFIELD: Learning 3D Feature Fields for Part Segmentation and Beyond},
  author={Liu, Minghua and Uy, Mikaela Angelina and Xiang, Donglai and Su, Hao and Fidler, Sanja and Sharp, Nicholas and Gao, Jun},
  journal={arXiv preprint arXiv:2504.11451},
  year={2025}
}

@article{liu2025freemesh,
  title={FreeMesh: Boosting Mesh Generation with Coordinates Merging},
  author={Liu, Jian and Weng, Haohan and Lei, Biwen and Yang, Xianghui and Zhao, Zibo and Chen, Zhuo and Guo, Song and Han, Tao and Guo, Chunchao},
  journal={arXiv preprint arXiv:2505.13573},
  year={2025}
}

@article{wang2025nautilus,
  title={Nautilus: Locality-aware Autoencoder for Scalable Mesh Generation},
  author={Wang, Yuxuan and Yi, Xuanyu and Weng, Haohan and Xu, Qingshan and Wei, Xiaokang and Yang, Xianghui and Guo, Chunchao and Chen, Long and Zhang, Hanwang},
  journal={arXiv preprint arXiv:2501.14317},
  year={2025}
}

@article{zhao2025deepmesh,
  title={DeepMesh: Auto-Regressive Artist-mesh Creation with Reinforcement Learning},
  author={Zhao, Ruowen and Ye, Junliang and Wang, Zhengyi and Liu, Guangce and Chen, Yiwen and Wang, Yikai and Zhu, Jun},
  journal={arXiv preprint arXiv:2503.15265},
  year={2025}
}

@article{tang2025efficient,
  title={Efficient Part-level 3D Object Generation via Dual Volume Packing},
  author={Tang, Jiaxiang and Lu, Ruijie and Li, Zhaoshuo and Hao, Zekun and Li, Xuan and Wei, Fangyin and Song, Shuran and Zeng, Gang and Liu, Ming-Yu and Lin, Tsung-Yi},
  journal={arXiv preprint arXiv:2506.09980},
  year={2025}
}

@article{lin2025partcrafter,
  title={PartCrafter: Structured 3D Mesh Generation via Compositional Latent Diffusion Transformers},
  author={Lin, Yuchen and Lin, Chenguo and Pan, Panwang and Yan, Honglei and Feng, Yiqiang and Mu, Yadong and Fragkiadaki, Katerina},
  journal={arXiv preprint arXiv:2506.05573},
  year={2025}
}

@article{deng2025geosam2,
  title={GeoSAM2: Unleashing the Power of SAM2 for 3D Part Segmentation},
  author={Deng, Ken and Yang, Yunhan and Sun, Jingxiang and Liu, Xihui and Liu, Yebin and Liang, Ding and Cao, Yan-Pei},
  journal={arXiv preprint arXiv:2508.14036},
  year={2025}
}

@article{zhao2025hunyuan3d,
  title={Hunyuan3d 2.0: Scaling diffusion models for high resolution textured 3d assets generation},
  author={Zhao, Zibo and Lai, Zeqiang and Lin, Qingxiang and Zhao, Yunfei and Liu, Haolin and Yang, Shuhui and Feng, Yifei and Yang, Mingxin and Zhang, Sheng and Yang, Xianghui and others},
  journal={arXiv preprint arXiv:2501.12202},
  year={2025}
}

@article{lei2025hunyuan3d,
  title={Hunyuan3d studio: End-to-end ai pipeline for game-ready 3d asset generation},
  author={Lei, Biwen and Li, Yang and Liu, Xinhai and Yang, Shuhui and Xu, Lixin and Huang, Jingwei and Tang, Ruining and Weng, Haohan and Liu, Jian and Xu, Jing and others},
  journal={arXiv preprint arXiv:2509.12815},
  year={2025}
}

@article{he2025materialmvp,
  title={MaterialMVP: Illumination-Invariant Material Generation via Multi-view PBR Diffusion},
  author={He, Zebin and Yang, Mingxin and Yang, Shuhui and Tang, Yixuan and Wang, Tao and Zhang, Kaihao and Chen, Guanying and Liu, Yuhong and Jiang, Jie and Guo, Chunchao and others},
  journal={arXiv preprint arXiv:2503.10289},
  year={2025}
}

@inproceedings{huang2025mv,
  title={Mv-adapter: Multi-view consistent image generation made easy},
  author={Huang, Zehuan and Guo, Yuan-Chen and Wang, Haoran and Yi, Ran and Ma, Lizhuang and Cao, Yan-Pei and Sheng, Lu},
  booktitle={ICCV},
  year={2025}
}

@article{hunyuan3d2025hunyuan3d,
  title={Hunyuan3D 2.1: From Images to High-Fidelity 3D Assets with Production-Ready PBR Material},
  author={Hunyuan3D, Team and Yang, Shuhui and Yang, Mingxin and Feng, Yifei and Huang, Xin and Zhang, Sheng and He, Zebin and Luo, Di and Liu, Haolin and Zhao, Yunfei and others},
  journal={arXiv preprint arXiv:2506.15442},
  year={2025}
}

@article{lai2025natex,
  title={NaTex: Seamless Texture Generation as Latent Color Diffusion},
  author={Lai, Zeqiang and Zhao, Yunfei and Zhao, Zibo and Yang, Xin and Huang, Xin and Huang, Jingwei and Yue, Xiangyu and Guo, Chunchao},
  journal={arXiv preprint arXiv:2511.16317},
  year={2025}
}

@article{chen2025lafite,
  title={LaFiTe: A Generative Latent Field for 3D Native Texturing},
  author={Chen, Chia-Hao and Zou, Zi-Xin and Cao, Yan-Pei and Yuan, Ze and Luo, Guan and Qi, Xiaojuan and Liang, Ding and Zhang, Song-Hai and Guo, Yuan-Chen},
  journal={arXiv preprint arXiv:2512.04786},
  year={2025}
}

@article{ma2025p3,
  title={P3-sam: Native 3d part segmentation},
  author={Ma, Changfeng and Li, Yang and Yan, Xinhao and Xu, Jiachen and Yang, Yunhan and Wang, Chunshi and Zhao, Zibo and Guo, Yanwen and Chen, Zhuo and Guo, Chunchao},
  journal={arXiv preprint arXiv:2509.06784},
  year={2025}
}

@article{zhao2025diception,
  title={Diception: A generalist diffusion model for visual perceptual tasks},
  author={Zhao, Canyu and Sun, Yanlong and Liu, Mingyu and Zheng, Huanyi and Zhu, Muzhi and Zhao, Zhiyue and Chen, Hao and He, Tong and Shen, Chunhua},
  journal={arXiv preprint arXiv:2502.17157},
  year={2025}
}

@inproceedings{le2025one,
  title={One diffusion to generate them all},
  author={Le, Duong H and Pham, Tuan and Lee, Sangho and Clark, Christopher and Kembhavi, Aniruddha and Mandt, Stephan and Krishna, Ranjay and Lu, Jiasen},
  booktitle={CVPR},
  year={2025}
}

@article{wu2025unilat3d,
  title={Unilat3d: Geometry-appearance unified latents for single-stage 3d generation},
  author={Wu, Guanjun and Fang, Jiemin and Yang, Chen and Li, Sikuang and Yi, Taoran and Lu, Jia and Zhou, Zanwei and Cen, Jiazhong and Xie, Lingxi and Zhang, Xiaopeng and others},
  journal={arXiv preprint arXiv:2509.25079},
  year={2025}
}

@article{he2025unipart,
  title={UniPart: Part-Level 3D Generation with Unified 3D Geom-Seg Latents},
  author={He, Xufan and Wu, Yushuang and Guo, Xiaoyang and Ye, Chongjie and Zhou, Jiaqing and Hu, Tianlei and Han, Xiaoguang and Du, Dong},
  journal={arXiv preprint arXiv:2512.09435},
  year={2025}
}

@article{lai2025lattice,
  title={LATTICE: Democratize High-Fidelity 3D Generation at Scale},
  author={Lai, Zeqiang and Zhao, Yunfei and Zhao, Zibo and Liu, Haolin and Lin, Qingxiang and Huang, Jingwei and Guo, Chunchao and Yue, Xiangyu},
  journal={arXiv preprint arXiv:2512.03052},
  year={2025}
}

@article{su2024roformer,
  title={Roformer: Enhanced transformer with rotary position embedding},
  author={Su, Jianlin and Ahmed, Murtadha and Lu, Yu and Pan, Shengfeng and Bo, Wen and Liu, Yunfeng},
  journal={Neurocomputing},
  volume={568},
  pages={127063},
  year={2024},
  publisher={Elsevier}
}

@inproceedings{esser2024scaling,
  title={Scaling rectified flow transformers for high-resolution image synthesis},
  author={Esser, Patrick and Kulal, Sumith and Blattmann, Andreas and Entezari, Rahim and M{\"u}ller, Jonas and Saini, Harry and Levi, Yam and Lorenz, Dominik and Sauer, Axel and Boesel, Frederic and others},
  booktitle={Forty-first international conference on machine learning},
  year={2024}
}

@article{zhou2024point,
  title={Point-sam: Promptable 3d segmentation model for point clouds},
  author={Zhou, Yuchen and Gu, Jiayuan and Chiang, Tung Yen and Xiang, Fanbo and Su, Hao},
  journal={arXiv preprint arXiv:2406.17741},
  year={2024}
}

@inproceedings{ma2025find,
  title={Find any part in 3d},
  author={Ma, Ziqi and Yue, Yisong and Gkioxari, Georgia},
  booktitle={ICCV},
  year={2025}
}

@article{zhu2025partsam,
  title={PartSAM: A Scalable Promptable Part Segmentation Model Trained on Native 3D Data},
  author={Zhu, Zhe and Wan, Le and Xu, Rui and Zhang, Yiheng and Chen, Honghua and Dou, Zhiyang and Lin, Cheng and Liu, Yuan and Wei, Mingqiang},
  journal={arXiv preprint arXiv:2509.21965},
  year={2025}
}

@article{paul2025name,
  title={Name That Part: 3D Part Segmentation and Naming},
  author={Paul, Soumava and Kaushik, Prakhar and Vaidya, Ankit and Bhattad, Anand and Yuille, Alan},
  journal={arXiv preprint arXiv:2512.18003},
  year={2025}
}

@article{ding2025fullpart,
  title={FullPart: Generating each 3D Part at Full Resolution},
  author={Ding, Lihe and Dong, Shaocong and Li, Yaokun and Gao, Chenjian and Chen, Xiao and Han, Rui and Kuang, Yihao and Zhang, Hong and Huang, Bo and Huang, Zhanpeng and others},
  journal={arXiv preprint arXiv:2510.26140},
  year={2025}
}

@inproceedings{dong2025one,
  title={From one to more: Contextual part latents for 3d generation},
  author={Dong, Shaocong and Ding, Lihe and Chen, Xiao and Li, Yaokun and Wang, Yuxin and Wang, Yucheng and Wang, Qi and Kim, Jaehyeok and Gao, Chenjian and Huang, Zhanpeng and others},
  booktitle={ICCV},
  year={2025}
}

@article{li2025moca,
  title={MoCA: Mixture-of-Components Attention for Scalable Compositional 3D Generation},
  author={Li, Zhiqi and Li, Wenhuan and Wang, Tengfei and Wang, Zhenwei and Wu, Junta and Wang, Haoyuan and Yang, Yunhan and Huang, Zehuan and Li, Yang and Liu, Peidong and others},
  journal={arXiv preprint arXiv:2512.07628},
  year={2025}
}

@article{yang2025partdiffuser,
  title={PartDiffuser: Part-wise 3D Mesh Generation via Discrete Diffusion},
  author={Yang, Yichen and Li, Hong and Zhu, Haodong and Yang, Linin and Lei, Guojun and Xu, Sheng and Zhang, Baochang},
  journal={arXiv preprint arXiv:2511.18801},
  year={2025}
}

@inproceedings{yang2025omnipart,
  title={Omnipart: Part-aware 3d generation with semantic decoupling and structural cohesion},
  author={Yang, Yunhan and Zhou, Yufan and Guo, Yuan-Chen and Zou, Zi-Xin and Huang, Yukun and Liu, Ying-Tian and Xu, Hao and Liang, Ding and Cao, Yan-Pei and Liu, Xihui},
  booktitle={Proceedings of the SIGGRAPH Asia 2025 Conference Papers},
  pages={1--12},
  year={2025}
}

@article{yan2025x,
  title={X-Part: high fidelity and structure coherent shape decomposition},
  author={Yan, Xinhao and Xu, Jiachen and Li, Yang and Ma, Changfeng and Yang, Yunhan and Wang, Chunshi and Zhao, Zibo and Lai, Zeqiang and Zhao, Yunfei and Chen, Zhuo and others},
  journal={arXiv preprint arXiv:2509.08643},
  year={2025}
}

@inproceedings{zeng2024paint3d,
  title={Paint3d: Paint anything 3d with lighting-less texture diffusion models},
  author={Zeng, Xianfang and Chen, Xin and Qi, Zhongqi and Liu, Wen and Zhao, Zibo and Wang, Zhibin and Fu, Bin and Liu, Yong and Yu, Gang},
  booktitle={CVPR},
  year={2024}
}

@inproceedings{liu2024texoct,
  title={Texoct: Generating textures of 3d models with octree-based diffusion},
  author={Liu, Jialun and Wu, Chenming and Liu, Xinqi and Liu, Xing and Wu, Jinbo and Peng, Haotian and Zhao, Chen and Feng, Haocheng and Liu, Jingtuo and Ding, Errui},
  booktitle={CVPR},
  year={2024}
}

@inproceedings{yu2023texture,
  title={Texture generation on 3d meshes with point-uv diffusion},
  author={Yu, Xin and Dai, Peng and Li, Wenbo and Ma, Lan and Liu, Zhengzhe and Qi, Xiaojuan},
  booktitle={ICCV},
  year={2023}
}

@article{yu2024texgen,
  title={Texgen: a generative diffusion model for mesh textures},
  author={Yu, Xin and Yuan, Ze and Guo, Yuan-Chen and Liu, Ying-Tian and Liu, Jianhui and Li, Yangguang and Cao, Yan-Pei and Liang, Ding and Qi, Xiaojuan},
  journal={ACM Transactions on Graphics (TOG)},
  volume={43},
  number={6},
  pages={1--14},
  year={2024},
  publisher={ACM New York, NY, USA}
}

@inproceedings{yuan2025seqtex,
  title={SeqTex: Generate Mesh Textures in Video Sequence},
  author={Yuan, Ze and Yu, Xin and Sun, Yangtian and Guo, Yuan-Chen and Cao, Yan-Pei and Liang, Ding and Qi, Xiaojuan},
  booktitle={Proceedings of the SIGGRAPH Asia 2025 Conference Papers},
  pages={1--12},
  year={2025}
}

@article{zhang2025bang,
author = {Zhang, Longwen and Zhang, Qixuan and Jiang, Haoran and Bai, Yinuo and Yang, Wei and Xu, Lan and Yu, Jingyi},
title = {BANG: Dividing 3D Assets via Generative Exploded Dynamics},
year = {2025},
issue_date = {August 2025},
publisher = {Association for Computing Machinery},
address = {New York, NY, USA},
volume = {44},
number = {4},
issn = {0730-0301},
url = {https://doi.org/10.1145/3730840},
doi = {10.1145/3730840},
journal = {ACM Trans. Graph.},
month = jul,
articleno = {62},
numpages = {21}
}

@article{ye2025hi3dgen,
  title={Hi3dgen: High-fidelity 3d geometry generation from images via normal bridging},
  author={Ye, Chongjie and Wu, Yushuang and Lu, Ziteng and Chang, Jiahao and Guo, Xiaoyang and Zhou, Jiaqing and Zhao, Hao and Han, Xiaoguang},
  journal={arXiv preprint arXiv:2503.22236},
  year={2025}
}

@article{wu2025direct3d,
  title={Direct3d-s2: Gigascale 3d generation made easy with spatial sparse attention},
  author={Wu, Shuang and Lin, Youtian and Zhang, Feihu and Zeng, Yifei and Yang, Yikang and Bao, Yajie and Qian, Jiachen and Zhu, Siyu and Cao, Xun and Torr, Philip and others},
  journal={arXiv preprint arXiv:2505.17412},
  year={2025}
}

@inproceedings{xue2023ulip,
  title={Ulip: Learning a unified representation of language, images, and point clouds for 3d understanding},
  author={Xue, Le and Gao, Mingfei and Xing, Chen and Mart{\'\i}n-Mart{\'\i}n, Roberto and Wu, Jiajun and Xiong, Caiming and Xu, Ran and Niebles, Juan Carlos and Savarese, Silvio},
  booktitle={CVPR},
  year={2023}
}

@article{zhou2023uni3d,
  title={Uni3d: Exploring unified 3d representation at scale},
  author={Zhou, Junsheng and Wang, Jinsheng and Ma, Baorui and Liu, Yu-Shen and Huang, Tiejun and Wang, Xinlong},
  journal={arXiv preprint arXiv:2310.06773},
  year={2023}
}

@inproceedings{feng2025romantex,
  title={Romantex: Decoupling 3d-aware rotary positional embedded multi-attention network for texture synthesis},
  author={Feng, Yifei and Yang, Mingxin and Yang, Shuhui and Zhang, Sheng and Yu, Jiaao and Zhao, Zibo and Liu, Yuhong and Jiang, Jie and Guo, Chunchao},
  booktitle={ICCV},
  year={2025}
}

@article{xiang2025native,
  title={Native and Compact Structured Latents for 3D Generation},
  author={Xiang, Jianfeng and Chen, Xiaoxue and Xu, Sicheng and Wang, Ruicheng and Lv, Zelong and Deng, Yu and Zhu, Hongyuan and Dong, Yue and Zhao, Hao and Yuan, Nicholas Jing and others},
  journal={arXiv preprint arXiv:2512.14692},
  year={2025}
}

@inproceedings{chen20253dtopia,
  title={3dtopia-xl: Scaling high-quality 3d asset generation via primitive diffusion},
  author={Chen, Zhaoxi and Tang, Jiaxiang and Dong, Yuhao and Cao, Ziang and Hong, Fangzhou and Lan, Yushi and Wang, Tengfei and Xie, Haozhe and Wu, Tong and Saito, Shunsuke and others},
  booktitle={CVPR},
  year={2025}
}

@inproceedings{ryu2025elevating,
  title={Elevating 3D Models: High-Quality Texture and Geometry Refinement from a Low-Quality Model},
  author={Ryu, Nuri and Won, Jiyun and Son, Jooeun and Gong, Minsu and Lee, Joo-Haeng and Cho, Sunghyun},
  booktitle={Proceedings of the Special Interest Group on Computer Graphics and Interactive Techniques Conference Conference Papers},
  year={2025}
}

@article{li2026segvigen,
  title={SegviGen: Repurposing 3D Generative Model for Part Segmentation},
  author={Li, Lin and Feng, Haoran and Huang, Zehuan and Chen, Haohua and Nie, Wenbo and Hou, Shaohua and Fan, Keqing and Hu, Pan and Wang, Sheng and Li, Buyu and others},
  journal={ACM Transactions on Graphics (TOG)},
  volume={45},
  number={4},
  pages={1--12},
  year={2026},
  publisher={ACM, New York, NY, USA}
}

@article{jin2025uniart,
  title={UniArt: Unified 3D Representation for Generating 3D Articulated Objects with Open-Set Articulation},
  author={Jin, Bu and Li, Weize and Gu, Songen and Zheng, Yupeng and Zheng, Yuhang and Zhou, Zhengyi and Yao, Yao},
  journal={arXiv preprint arXiv:2511.21887},
  year={2025}
}

@article{zhao2026flexpart,
  title={FlexPart: Text-Controlled Semantics-Aware 3D Part Segmentation at Flexible Granularity},
  author={Zhao, Yiqun and Xu, Jiachen and Zhi, Yihao and Zhao, Yunfei and Zhao, Zibo and Han, Xintong and Guo, Chunchao and Gao, Shenghua},
  journal={arXiv preprint arXiv:2610.23659},
  year={2026}
}

@article{tsui2026point2part,
  title={Point2Part: Unified 3D Partitioning from Point Prompts},
  author={Tsui, Hao-Tang and Tuan, Yu-Rou and Ma, Xiaoxuan and Ugrinovic, Nicolas and Shiratori, Takaaki and Kitani, Kris},
  journal={arXiv preprint arXiv:2609.38180},
  year={2026}
}

@article{zhao2026seeing,
  title={Seeing Both the Forest and the Trees: Reusing Holistic 3D Priors for Part-Decomposed Generation},
  author={Zhao, Yiqun and Huang, Binbin and Duan, Haobin and Zhao, Zibo and Gao, Shenghua},
  journal={arXiv preprint arXiv:2610.21850},
  year={2026}
}

@article{zhu2026partllm,
  title={PartLLM: A Unified Multimodal Foundation for 3D Part Segmentation},
  author={Zhu, Zhe and Zhang, Yiheng and Li, Peng and Zhao, Zixing and Chen, Honghua and Zhang, Yaqing and Wan, Le and Dou, Zhiyang and Lin, Cheng and Liu, Yuan and others},
  journal={arXiv preprint arXiv:2609.25832},
  year={2026}
}

@article{gu2026unified,
  title={A Unified Framework for Image-to-3D Part Generation via Variable Granularity},
  author={Gu, Qiwen and Zhou, Xin and Chen, Xiwu and Zhu, Zhongyang and Zhang, Zhuo and Zhao, Junqiao},
  journal={arXiv preprint arXiv:2610.31671},
  year={2026}
}

@article{lin2026seg3dparts,
  title={Seg3DParts: Segmentation-Grounded Controllable Part-Level 3D Generation},
  author={Lin, Jiantao and Chen, Meixi and Xu, Yingjie and Fu, Chenbo and Wu, Leyi and Chen, Hao and Li, Yinchuan and Chen, Ying-Cong},
  journal={arXiv preprint arXiv:2609.36918},
  year={2026}
}

@inproceedings{su2026s,
  title={{S$^2$AM3D}: Scale-controllable Part Segmentation of 3D Point Clouds},
  author={Su, Han and Huang, Tianyu and Wan, Zichen and Wu, Xiaohe and Zuo, Wangmeng},
  booktitle={CVPR},
  year={2026}
}

@article{hadgi2026patchalign3d,
  title={Patchalign3d: Local feature alignment for dense 3d shape understanding},
  author={Hadgi, Souhail and Gong, Bingchen and Sundararaman, Ramana and Pierson, Emery and Li, Lei and Wonka, Peter and Ovsjanikov, Maks},
  journal={arXiv preprint arXiv:2601.02457},
  year={2026}
}

@article{jin2026cosmo3d,
  title={CoSMo3D: Open-World Promptable 3D Semantic Part Segmentation through LLM-Guided Canonical Spatial Modeling},
  author={Jin, Li and Chen, Weikai and Wang, Yujie and Yin, Yingda and Hu, Zeyu and Zhang, Runze and Luo, Keyang and Qian, Shengju and Wang, Xin and Qin, Xueying},
  journal={arXiv preprint arXiv:2603.01205},
  year={2026}
}

\end{document}